\documentclass[preprint,review,11.5pt]{elsarticle}
\usepackage[margin=1in]{geometry}
\usepackage{caption}
\usepackage{subcaption}
\usepackage{amsmath}
\usepackage{amsfonts}
\usepackage{xcolor}
\usepackage{soul}
\usepackage{hyperref}
\usepackage{tabularx}
\usepackage{enumitem}
\usepackage{float}

\journal{arXiv}

\begin{document}

  






\begin{frontmatter}


\author{Mohsen Jozani\fnref{label1}}

\author{Jason A. Williams\corref{cor1}\fnref{label1}}
 \ead{jwilliams45@augusta.edu}
\cortext[cor1]{}

\author{Ahmed Aleroud\fnref{label1}}
\author{Sarbottam Bhagat\fnref{label2}}
 
\title{The Role of Emotions in Informational Support Question-Response Pairs in Online Health Communities: A Multimodal Deep Learning Approach}

 \affiliation[label1]{organization={Augusta University School of Computer and Cyber Sciences},
             addressline={1120 15th Street},
             city={Auguta},
             state={GA},
             postcode={30192},
             country={USA}}
\affiliation[label2]{organization={University of Wisconsin Eau-Claire College of Business},
             addressline={1702 Park Ave},
             city={Eau-Claire},
             state={WI},
             postcode={54701},
             country={USA}}



\begin{abstract}
This study explores the relationship between informational support seeking questions, responses, and helpfulness ratings in online health communities. We created a labeled data set of question-response pairs and developed multimodal machine learning and deep learning models to reliably predict informational support questions and responses. We employed explainable AI to reveal the emotions embedded in informational support exchanges, demonstrating the importance of emotion in providing informational support. This complex interplay between emotional and informational support has not been previously researched. The study refines social support theory and lays the groundwork for the development of user decision aids. Further implications are discussed. 

\end{abstract}


\end{frontmatter}


\section{Introduction}
\vspace{-2mm}
People who care for their own health or the health of their loved ones often seek support in online health communities (OHCs). OHCs provide a valuable resource to people who do not have access to healthcare services or to those who cannot afford those services \cite{Chen2019}\cite{Yin2023Bundled}. Additionally, OHCs serve as a platform for patients to ask questions they may feel hesitant to ask their healthcare providers \cite{Gui2017}. Thus, OHCs offer a digital complement or alternative to in-person social support exchanges. Social support is defined as an exchange of resources between individuals, where one or both parties perceive the fundamental need of the recipient as being met \cite{shumaker1984toward}. While social support in an offline setting can take on various forms, emotional support and informational support are the major types of support found in OHCs \cite{Chen2019, braithwaite1999communication, mo2008exploring, huang2014social, coulson2007social}. Emotional support involves exchanging empathy, care, and encouragement with the goal of reducing negative feelings and avoiding anxiety or depression \cite{Zhou2023}. Informational support refers to the request or provision of actionable information that can reduce uncertainty and help in solving problems \cite{yan2014feeling, huang2019sharing, Park2020Disentangling}. Past research has significantly improved our understanding of support seeking \cite{chen2020linguistic} and support provisioning \cite{yan2014feeling, Zhou2023} behaviors of OHC users. However, there are certain areas that demand closer attention.

First, it is important to ensure that the type of support provided in responses aligns with the type of support requested by the questioner. This is particularly important for informational support requests, as they account for the majority of questions on OHCs \cite{yan2014feeling}. Requests for information should be met with the desired information. Informational support responses, also more prevalent and common than emotional support responses \cite{yan2014feeling}, reduce the information asymmetry between the person asking the question and the respondent \cite{chen2020linguistic}, and in doing so they can effectively address the question or concern of the information seeker \cite{Wyer1999}. 

Second, while past research has often treated and predicted emotional support and informational support as mutually exclusive outcomes, the literature acknowledges their coexistence \cite{yan2014feeling}. A request for informational support is an attempt to reduce the uncertainty that is causing fear, sadness, and anxiety \cite{brashers2004social}. Therefore, one would expect to observe the indicators of these emotions in the question of the information seeker. And, just as there are linguistic similarities between questions and responses in OHCs \cite{chen2020linguistic}, the emotions embedded in an informational support seeking question may also be mirrored in the responses. Additionally, emotional support requests may sometimes be concealed within informational seeking questions \cite{bambina2007online}. Consequently, it is essential to consider the emotional valence of text in determining informational support questions and responses. 

Third, while previous studies have explored the types of support provided in OHCs and their effects, there has been limited investigation into the specific relationship between informational support responses (ISRs) and their perceived helpfulness. OHCs serve as vital platforms for individuals seeking guidance, advice, or solutions to their health-related concerns. Understanding whether ISRs predict the helpfulness of responses is a crucial inquiry within the context of OHCs. By doing so, one may discern what aspects of ISRs have the most significant impact on the helpfulness of such support responses. Furthermore, OHCs consist of diverse sub-communities, each focusing on specific medical conditions or topics. These sub-communities often have unique norms, preferences, and expectations regarding the helpfulness of responses \cite{ronghua2017research}. Investigating the relationship between ISRs and response helpfulness may be pivotal for obtaining a more comprehensive understanding of support dynamics within OHCs.

Fourth, past research has mainly used statistical methods and conventional machine learning (ML) techniques to classify social support types. The advent of large language models in recent years has revolutionized natural language processing (NLP). Social support research can benefit from these advanced models to achieve improved classification accuracy and enhanced model transferability. We show that our model exhibits the potential for transferability to new sub-communities, enabling accurate ISR predictions with limited labeled data. To put this concept to the test, we evaluate the  model adaptability to different sub-communities each with a different medical condition and show that our model has good transferability across them.

This study aims to explore these relationships between informational support seeking questions (ISSQ) and ISRs in OHCs with a focus on explaining the role of emotions in both the questions and responses, and examining whether ISR predict response helpfulness. Specifically, this study seeks to answer the following research questions:

\begin{enumerate}[noitemsep]
    \item How does the pairwise consideration of ISSQs and ISRs enable reliable prediction of informational support in OHCs and transfer this predictive capability to other sub-communities, each representing distinct health conditions in OHCs?
    \item What are the roles of valenced and neutral emotions in determining informational support in questions and responses?
    \item Do informational support responses predict the helpfulness of a response?
\end{enumerate}

To achieve these goals, we employ language models and a multimodal deep learning (DL) approach to predict the solicitation and provision of informational support in OHCs. Driven by the social support framework \cite{yan2014feeling, sarason2013social} and inspired by the coding scheme developed in \cite{bambina2007online}, we labeled a random sample of 2,000 question-response pairs taken from four types of medical conditions in an OHC. Then, we developed ML and DL models to classify questions as informational or emotional support seeking. Next, we trained a second set of models to identify ISRs. Our approach involves combining state-of-the-art language models with the predictive capacity of emotional features and other numerical features identified in prior research to develop high performing models. We demonstrated the generalizability of our findings by applying our best performing models to another sub-communities, each representing distinct health conditions. Next, we used explainable AI techniques to investigate the impact of emotions and other numerical features such as text metadata, post-related and user-related attributes on our model predictions. Lastly, logistic regression was used to examine the effect of ISR on response helpfulness.

It is thought that emotions would not play a significant part in ISRs \cite{bambina2007online}. However, we provide evidence that emotions in informational responses are not absent from ISRs and are valuable at predicting informational support. Our further examination reveals that informational support serves as a strong and positive predictor of perceived helpfulness of a response. We contribute to social support theory by refining the definition of informational support. We also contribute to NLP research in IS by demonstrating state of the art ML techniques. This research can also lead to better medical decision aids and emotion-sensitive AI chatbots.

The rest of this paper is organized as follows. The literature review is discussed in Section 2. Section 3 describes the data, features, and methodology. Section 4 reports the analysis results. Section 5 discusses the findings, theoretical and practical contributions. Section 6 details the study limitations, and future research directions.

\section{Literature Review}
\vspace{-2mm}
\subsection{Social Support Theory}
This study draws upon social support theory \cite{sarason2013social} to provide a comprehensive framework for investigating the relationship between ISSQs and ISRs within OHCs. Central to the fabric of human interactions, social support theory posits that individuals engage in a reciprocal exchange of resources within social networks to address perceived needs and enhance their own well-being \cite{Dwivedi2023SocialCommerce}. Social support fulfills fundamental needs of individuals through interactions that can be emotional, informational, or instrumental in nature. Social support theory is adeptly applied to the specific research context of OHCs, where individuals seek and provide support through text-based interactions. The theory's concepts illuminate the multifaceted nature of support-seeking behaviors, where individuals may not only require factual information but also emotional reassurance and connection. 

\subsection{Informational Support Questions and Responses}
Informational support has been defined as the provision of information, advice, and guidance to individuals to help them address a specific issue or problem with an aim to offer knowledge, perspectives or insights to help them make informed decisions or take appropriate actions \cite{sarason2013social, krause1986social, langford1997social, wills1991social}. In the context of OHCs, ISRs refer to specific responses that aim to provide individuals seeking health-related information or support with accurate and useful information, advice, or resources. An ISR informs the question-asking user about their specific medical question(s) by providing actionable, relevant and useful information ranging from symptoms, diagnosis, treatment options, risks, and benefits associated with medications and surgeries, as well as preventative measures whenever applicable. These responses may be provided by healthcare professionals, peers, or moderators of the online health community and are intended to help individual make informed decisions about their health and well-being by conveying the knowledge that the person asking the question does not already have \cite{Wyer1999}, reducing information asymmetries \cite{chen2020linguistic}. 

This characteristic of ISRs assumes that the person asking the question can read and comprehend the response \cite{chen2020linguistic}. This means the value of a given response could only be determined if such a response adequately ``informs” the user in ways that align with their expectations for communications. However, the current understanding of what makes responses rich in informational support is limited in the context of OHCs due to the complexity and diversity of health-related questions, the constantly evolving nature of OHCs, and the variability in quality of responses influenced by contextual factors and social dynamics. 

\subsection{Informational Support and Information Helpfulness}
In the context of OHCs, informational support plays a crucial role in providing individuals seeking health-related information or support with accurate, helpful and useful information, advice, or resources \cite{bhagat2022conceptualizing}. The definition of informational support draws from established literature on social support (e.g., \cite{sarason2013social, krause1986social, langford1997social, wills1991social}), including the provision of information, advice, and guidance to help individuals address specific issues or problems. This form of social support aims to offer knowledge, perspectives, or insights that empower individuals to make informed decisions or take appropriate actions concerning their health and well-being \cite{Wyer1999}.

According to this definition, ISRs encompass the idea of helpfulness, making it inherently intertwined with the concept of informativeness. In other words, something cannot be deemed ``informational” within the context of OHCs if it does not possess inherent utility, helpfulness, or usefulness. This perspective aligns with the fundamental purpose of ISRs, which is to inform the question-asking user in a manner that addresses their specific medical questions and provides actionable, relevant, and practical information. Such information may cover a wide spectrum, including symptoms, diagnoses, treatment options, risks and benefits associated with medical interventions, and preventative measures when applicable. However, it is crucial to acknowledge that the value of an ISR can only be determined when it adequately ``informs" the user in ways that align with their expectations for communication. This consideration is particularly important in the context of OHCs, where health-related questions are diverse, and the quality of responses may be influenced by contextual factors and social dynamics because these factors play a pivotal role in shaping users' perceptions of the information they receive, which, in turn, impacts their trust in the community and the decisions they make regarding their health \cite{daraz2019can, bhagat2022conceptualizing}.

\subsection{Combining Informational and Emotional Support}


 Recent studies indicate that in some cases, patients may express concerns about the clarity and relevance of information received from their healthcare professionals \cite{wang2021online}. While healthcare professionals play a vital role in providing medical guidance, it is important to acknowledge that patient perceptions of information can vary depending on individual circumstances. Patients seek information from OHCs to fulfill their unmet needs for health information \cite{pendry2015individual, prescott2019young}. Users may also rely on OHCs to share various emotional support requests from other users \cite{Peng2020}. These requests are often in the form of health updates, venting frustrations, and requests for prayer. For example, an examination of user participation in an online health community in China found that emotional and informational support were positively associated with participation \cite{wang2017analyzing}. 

Informational, emotional, and other forms of social support have been studied in the context of OHCs \cite{coulson2007social, huang2014social}. These studies found that the extent of social interaction and social identification with the community determined the provision of emotional support, while healthcare-related expertise predicted informational support provision. A study focusing on the associations between OHC use and users' well-being found that social support provided by OHCs was positively associated with patients' emotional well-being. It further found that perceived quality of support, including emotional and informational support, was more important than frequency of support exchanges in predicting well-being outcomes \cite{yan2014feeling}.
Another study looked into the influence of risk perception attitude and social support on health information-seeking behavior on mobile social media platforms \cite{deng2017understanding}. The study found that tangible and appraisal support significantly influenced perceived risk, while emotional and esteem support significantly influenced health self-efficacy.


With both informational and emotional support being common on OHCs and appearing in the same support threads, it is reasonable to expect them to also appear in the same message
\cite{jong2021exchange, solberg2014benefits}. Likewise, 
affective and informative linguistic signals are effective in invoking both informational and emotional support from the community \cite{chen2020linguistic}. Prior research has examined emotional and informational elements in online support groups for individuals with chronic health conditions and found that both emotional and informational support were present in the messages posted by participants, with emotional support being more prevalent in the early stages of group formation and informational support increasing over time \cite{deetjen2016informational}. Analyzing the linguistic features of OHC messages, a study revealed that longer queries often related to eliciting informational support responses. Additionally, the inclusion of negative sentiments in a question amplifies the likelihood of receiving both informational and emotional support responses \cite{jiang2022effect}.

These studies provide insights into the social support dynamics within OHCs. While these studies have all found both types of support in OHCs, they have never considered the emotional content of informational support. That is, informational support delivered in its most natural form: from one person trying to help another, using emotional language as needed to convey meaning.
Social support theory allows us to delve into such nuanced interactions observed within OHCs, offering a robust framework for examining the exchange of resources and provision of support among individuals, specifically focusing on how support interactions contribute to the overall well-being of participants. 

The present study emphasizes the prediction of informational support provided by OHC users through the integration of emotions (valenced and neutral), and text-based, response-related, and user-related features. Additionally, we explored the link between the provision of information and the users' perception of the helpfulness of a response. Our analysis fully incorporates the intertwined nature of emotional and informational support, as understood in social support theory. This integration enables us to unravel hidden emotional dimensions within informational support interactions, thereby deepening our understanding of the underlying motivations and sentiments of participants. The application of this theory to our study enriches our understanding of how emotional states drive individuals to seek information and how the provision of information can, in turn, affect perceived helpfulness. Such insights enable us to address the central research questions more comprehensively, unveiling the intricate connections between emotions and information-seeking behaviors.

\subsection{Machine Learning Techniques in OHCs}

Several studies have used ML methods to study OHCs. One recent study applying ML to OHCs has found that linguistic signals play a significant role in influencing emotional and informational support \cite{chen2020linguistic}. These signals include negative sentiments, readability, linguistic style matching, spelling, and post length. Using a support vector machine (SVM) classifier, the study revealed positive correlations between negative sentiments and readability with both types of support. Additionally, linguistic style and convergent post content were linked to higher support levels, while longer posts showed a positive correlation with informational support and a negative correlation with emotional support. 

Wang and colleagues conducted a limited set of experiments to classify messages as providing emotional or informational support using linguistic inquiry and word count (LIWC), linguistic features, and latent Dirichlet allocation (LDA) topical features, yielding an accuracy of 78\% \cite{wang2012stay}. Notably, emotional support had a significant impact on users' intention to leave forums compared to informational support, with community characteristics playing a role in this correlation.

In another study, a simple Bayesian ML model was used to examine support-seeking behaviors, revealing that mental health conditions tended to elicit emotional support, whereas physical medical condition such as diabetes prompted informational support \cite{deetjen2016informational}. Furthermore, the study identified differences in support-seeking patterns among gender and age groups, along with limitations related to their use of a bag-of-words model. In our current research, we address this limitation through the use of advanced NLP models and a multimodal ML approach.

Another related work conducted ML-based experiments to identify different types of social support (i.e., seeking and providing informational support and emotional support) \cite{wang2017analyzing}. The study identified informational support as the most sought-after form of social support, whereas seeking and receiving informational support were negatively linked to participation. This research predominantly used numerical features to quantify the number of support posts and employed conventional ML techniques such as naive Bayesian, random forest, and decision tree.
\vspace{-3mm}


\section{Methods}
\begin{figure}
    \centering
\includegraphics[width=1\columnwidth]{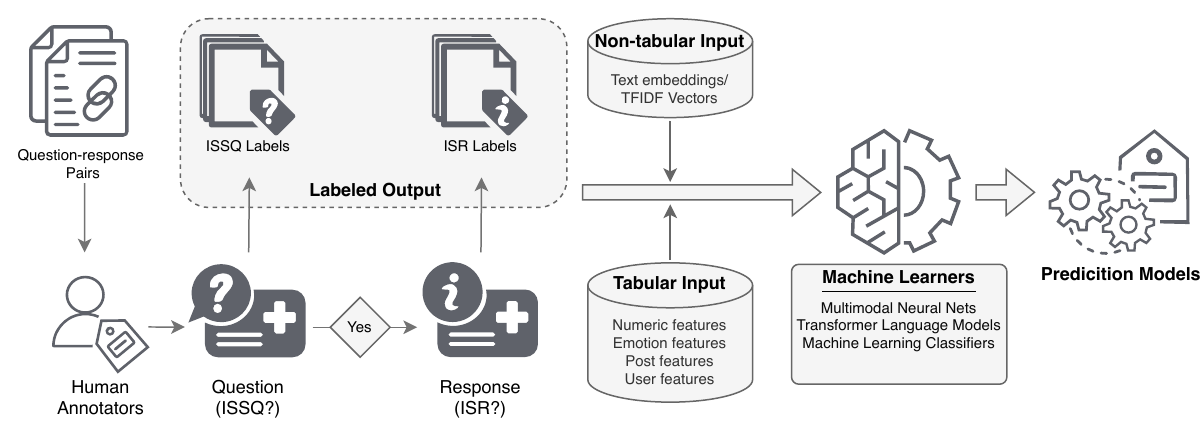}
    \caption{Research Process}
    \label{fig:my_label}
\end{figure}

\vspace{-2mm}
The overall research method is summarized in Figure \ref{fig:my_label}. 2,000 question-response pairs from four types of medical conditions were randomly selected. Each question response pair was separately labeled by three human raters. Next, textual features were identified for each question and response. These were fed into several different classification models. 

\subsection{Data Collection}
Data was obtained from MedHelp (medhelp.com), a popular OHC. We collected the content of over 1.1 million questions, including 6 million responses to those questions, and the profile information of the 15 million users of the platform. At the time of data collection, MedHelp had 270 sub-communities, each representing a unique medical condition. We excluded non-English sub-communities and those that were not directly related to human health (such as the ones on dogs, parenting, and relationships), reducing the number of sub-communities to 137. From there, we focused on the following medical conditions that had several posts in each of their sub-communities: cancer, diabetes, neurological issues, and cardiovascular problems. Sub-communities related to these medical conditions displayed a large variance in the number of posts they contain. 

\subsection{Labeling Process}
To determine an appropriate sample size for data labeling, we reviewed previous studies (e.g., \cite{Zhou2023, chen2020linguistic, wang2017analyzing}) and concluded that labeling 2,000 randomly-selected question-response pairs would be sufficient. We focused on four major medical conditions: cancer, diabetes, cardiovascular, and neurological conditions. We used stratified random sampling to create a semi-balanced data set that contains approximately equal observations within the sub-communities. Following the coding scheme in \cite{bambina2007online, yan2014feeling, radin2006me}, we developed a guideline (Appendix A) to establish what qualifies as an ISSQ and what constitutes an ISR response.

\begin{table}[h]
\centering
\caption{Labeling Guidelines}
\begin{tabular}{p{16cm}}
\hline
\small
\textbf{Labeling Informational Support Seeking Questions (ISSQs) (adapted from \cite{yan2014feeling})} 
\vspace{-2mm}
\begin{itemize}[noitemsep]
     \item The post contains or implies a specific medical question, about the pains, treatments, procedures, diagnosis, and lifestyle habits associated with it.   
     \item The questioner solicits an actual response, advice, or guidance from the community.
    \item Excludes vent-offs, health updates, community updates, and news content sharing.

\end{itemize}
\textbf{Labeling Informational Support Seeking Responses (ISRs) (adapted from \cite{bambina2007online})}
\vspace{-2mm}
\begin{itemize}[noitemsep]
    \item Answers the ISSQ by providing advice, guidance, or personal experience that can help in solving the questioner's problem.
        \subitem - Responses to see a physician are not informational as they do not reduce the information asymmetry, do not directly answer the question, and provide no meaningful course of action that can be undertaken without further input from a physician.
        \subitem - Referrals to other users, posts, or websites are not informational as they do not directly answer the question.
\end{itemize}
\\\hline
\end{tabular}
\end{table}

Using the guidelines in Appendix A, three raters individually evaluated each question to ensure that the questioner was seeking informational support instead of pursuing other purposes such as seeking emotional support \cite{Peng2020}. If a question was determined to be an ISSQ, the raters also examined if the response was an ISR. After the first round of rating, there was 91\% and 79\% agreement on what constituted ISSQs and ISRs, respectively. After discussion to address the discrepancies in evaluations, the raters achieved 97.8\% and 97.2\% agreement on the ISSQ and ISR labeling tasks, respectively.

Table \ref{tab:LabelExamples} gives examples of several question-response pairs in the data. The question in row 1 is not an ISSQ because it does not ask a medical question. The question in row 2 is an ISSQ because the questioner lays out their symptoms and ask about a test to further the diagnosis. The response in row 2 is not an ISR because it does not answer the ISSQ. The question in row 3 is an ISSQ because it seeks advice on getting pain relief from a headache. The response in row 3 is an ISR because even though it recommends a vision test (a physician referral) it also provides several possible courses of action for the questioner to take in lieu of a physician visit.

\begin{table}[t!]
\footnotesize
    \caption{Examples of Question-Response Pairs and Labels}
    \label{tab:LabelExamples}
\begin{tabularx}{\textwidth}{XXll}
\hline
\textbf{Question}          & \textbf{Response} & \textbf{ISSQ} & \textbf{ISR} \\ \hline
Whats on everyone's agenda for this week?
& I go tomorrow to apply for some disability. I don't know when I'll be able to get to a doctor to schedule a TTT or other tests I think I need. Aside from that, I've been really sick the past 2 days or so and woke up today feeling drained. Sleep doesn't seem to refresh me lately. & No & No \\ \hline
I have been feeling light-headed from time to time very frequently. I cannot tilt my head back, or else I would feel very dizzy and nauseated.  My vision is also getting worse. Things are getting more blurry now.  I do get frequent tension and migraine headaches. My neck area feels stiff and sore.  My memory is getting worse too. I seem to forget things easily and sometimes when I'm in the middle of doing something, I feel confused. Should I go get a brain scan? &Have you at least seen your doctor?
 & Yes & No \\ \hline
I'm been having headache every single day. Sometimes once a day, sometimes twice. With the headache accomplished with a rining in the ear or head, I also can hear my own heartbeat when I lay down. I have CT with contrast done. Nothing wrong. I have been taken alots of medication like dexdron, medrox. tylenol, Butalbital, and vitmanin B complex. What can I do to get rid of this headache? & Hi, I would also suggest you to for vision test for eyesight. Also taking too many pain medications can cause rebound headache so avoid taking too many medications at the same time. Are you getting a good nights sleep? I would also suggest you eat your meals on time to prevent hypoglycaemia which can cause headache.Sleep well, eat on time and maintain your fluid intake and see if all this makes any difference to your headache. Good luck! & Yes & Yes \\ \hline
\end{tabularx}
\end{table}

\subsection{Features}
Past research has identified a list of textual, affective, and psychometric features to distinguish between the types of social support that may exist within health messages \cite{chen2020linguistic, yan2014feeling, Zhou2023}. These studies have used a range of approaches from bag-of-words probabilistic classifiers \cite{yan2014feeling} to supervised ML approaches including SVM and tree-based techniques \cite{chen2020linguistic, wang2017analyzing}. However, only recently scholars have started to use advanced neural networks to identify different types of social support. Zhou et al. \cite{Zhou2023} used a compound hierarchical attention network model (C-HAN) to concatenate the text from the original post with post replies and sequences. Their best performing model achieved an overall accuracy of 84.7\% in distinguishing ``Emotional Support” from ``Auxiliary Content” which contained informational support. 

Transformer-based language models are the current industry standard for NLP and have achieved exceptional performance in a variety of text classification tasks \cite{lee2020biobert}. However, they have not been used in social support literature. Besides, prior studies have either used non-tabular word representations (the document itself), or tabular features computed and extracted from the text (e.g., text length, readability, and grammatical score). We seek to address these methodological gaps by leveraging the predictive capabilities of language models and use them along with emotional scores and the numerical features identified by the prior works using a multimodal fusion approach. We rely on language model embeddings and Term Frequency - Inverse Document Frequency (TF-IDF) vectors for our non-tabular text data and use several ML, DL, and ensemble approaches to account for tabular features including negative and positive emotions, numeric attributes such as text length, number of running statements and queries, question-response cosine similarity scores, as well as the post related and user related attributes.
The features examined are discussed below:

\subsubsection{Non-tabular Features}
Non-tabular data refers to unstructured data such as text, image, or video that cannot be formatted and represented in a table. We consider two types of word representations.

\noindent \textbf{Term Frequency - Inverse Document Frequency (TF-IDF) - } it was employed to predict ISR with the conventional ML algorithms. TF-IDF creates large vectors with rows representing documents and columns representing unique words. The consideration of inverse term frequency means that higher scores are given to words that rarely appear in the entire corpus but frequently in a single document and lower scores to words that appear frequently in all documents.

\noindent \textbf{Text Embeddings - } for approaches involving language models, instead of TF-IDF we use text embedding techniques that convert words into continuous vector spaces where similar or related words have similar representations. Large language models (such as GPT, BERT, and ELECTRA) use deeply contextualized embeddings that are trained on massive textual data and can be fine-tuned for downstream regression or classification tasks \cite{devlin2018bert, Wililams2023Detecting}.

\subsubsection{Tabular Features}
Tabular features include numerical, categorical, and binary values that can be represented in a traditional column-row format. We consider emotions along with other numeric features identified in the past research. We validate the effect of these features on our outcome variables using classical statistical models as shown in Appendix B.

\noindent \textbf{Emotions -} Ekman \cite{ekman1992argument} describes six fundamental emotions that serve an important evolutionary purpose, and therefore, are shared among people universally regardless of culture, region, or language. In this study, we use this framework to measure the intensity of valenced emotions of \textit{joy}, \textit{surprise}, \textit{sadness}, \textit{fear}, \textit{disgust}, and \textit{anger}. To account for texts that may not exhibit any noticeable emotion, we also compute a neutral emotion score.

\textit{Joy} is as a positive emotion that covers thrill, bliss, relief, happiness, and content. \textit{Surprise} pertains to the feeling of being in awe or shock from an unexpected news or an event. Depending on the trigger, surprise can either be a positive or a negative emotion \cite{ekman1985startle}. \textit{Sadness} refers to a sense of sorrow, hopelessness, and grief arising from a loss of someone or something important. \textit{Fear} describes being worried or anxious about an event or individual. It is a response to a perceived threat or source of danger. \textit{Disgust} is a feeling of aversion towards something that could be perceived as offensive or wrong. It may range from mild dislike to intense loathing. Lastly, \textit{anger} describes a feeling of rage, hostility, dislike, or disapproval towards a problem, event or an individual. The distributions of emotion classes for both questions and responses are shown in Appendix D.

To compute emotion scores, we utilize a publicly available Transformer-based model called DistilRoBERTa that had been fine-tuned on six distinct data sets specifically for this task \cite{hartmann2022emotionenglish}. This emotion prediction framework is suitable for our purpose as it has been trained on a diverse range of text types, including social media posts, self-reports, and TV dialogue utterances. Emotion scores are calculated as continuous values that range from 0 to 1.

\noindent \textbf{Text Metadata}

\noindent \textbf{Count of running statements (\small Q\_STATEMENT, R\_STATEMENT) -} the number of sentences in each text.

\noindent \textbf{Count of running queries (\small Q\_QUERY, R\_QUERY) -} the number of queries/questions in each text. 

\noindent The above two tasks are accomplished using a miniBERT pre-trained language model that can distinguish a query from a statement \cite{khan_2021}. We split each text into separate sentences, passed each sentence through the model and counted the number of query/statement sentences in each text.

\noindent \textbf{Text length (\small Q\_LEN, R\_LEN) -} represents the number of words in each question/response. Word count is considered the most popular measure of text length \cite{fleckenstein2020long}.

\noindent \textbf{TF-IDF cosine similarity (\small TFIDF\_CS) -} measures the similarity between words in each question-response pair. This score was computed by summing the TF-IDF weights of similar words in a response found in the question, adjusted by their cosine similarity scores \cite{riedel2017simple, ebadi2021memory}.

\noindent\textbf{Post Related Features}

\noindent \textbf{Medical Condition (\small M\_CONDITION) -} the name of the medical condition. This determines the type of medical conditions discussed.

\noindent \textbf{Response sequence (\small RID) -} the order in which the response was posted (self-developed).

\noindent \textbf{Questioner's engagement (\small Q\_REPLY\_RATIO) -} the count of replies made to the post by the questioner over the sum of all responses (self-developed).

\noindent \textbf{User Related Features}

\noindent \textbf{User's platform tenure (\small Q\_TENURE, R\_TENURE) -} the delta time (in seconds) between posting the response/ question and user's membership on the platform (self-developed).

\noindent \textbf{User's platform-wide response count (\small Q\_PWRC, R\_PWRC) -} the total number of responses posted by the user on the platform (self-developed).

\noindent \textbf{User's medical expertise (\small Q\_MED\_EXPERT, R\_MED\_EXPERT)-} user's self-disclosed profession (medical doctor, nurse, physician, specialist, etc.) coded as binary (self-developed).

\begin{figure}
    \centering
    \includegraphics[width=0.8\textwidth]{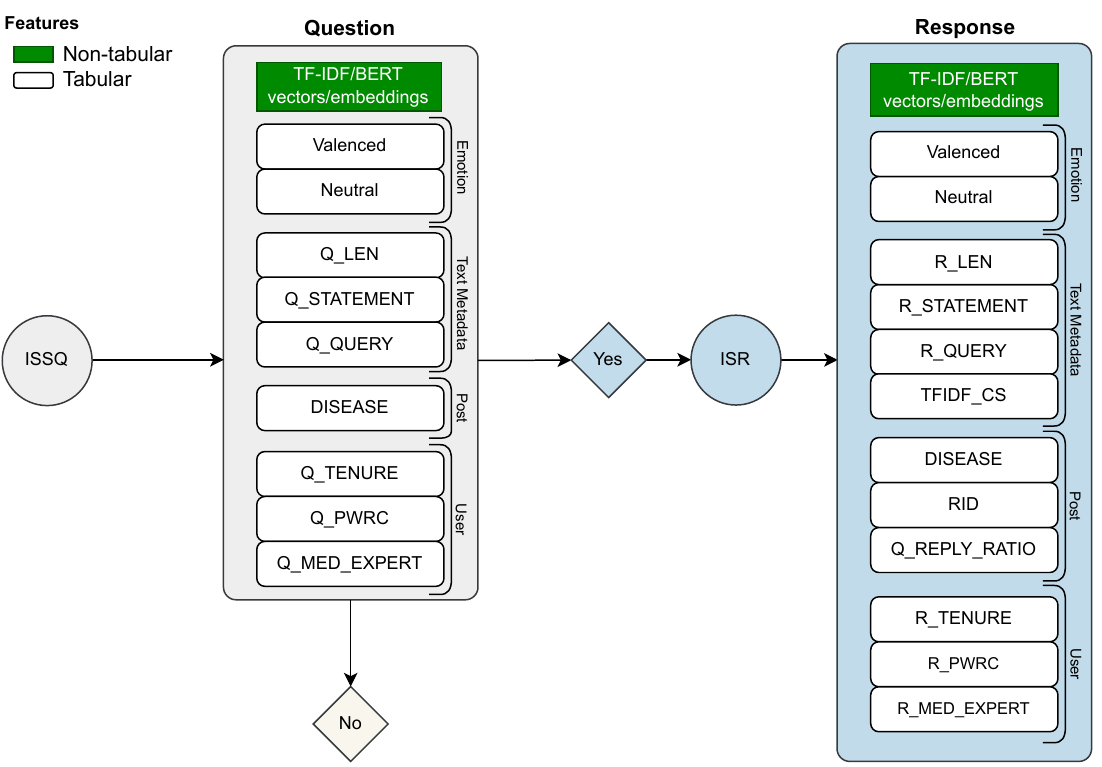}
    \caption{The features used in the process of label prediction.}
    \label{fig:features}
\end{figure}

The features used in completing the two tasks of identifying medical questions and informative responses are shown in Figure \ref{fig:features}. To measure perceived helpfulness for research question 3, we rely on whether a response has been marked as helpful (a binary designation) by the users of the platform.

\subsection{Data Modeling and Machine Learners}
We investigated the application of traditional ML approaches, including SVM, K-Nearest Neighbors (K-NN), Random Forest, and Logistic Regression for both classification tasks ISSQ and ISRs using tabular and/or text features. We also applied Gradient Boosting which is an ensemble ML technique that combines multiple learners to create a stronger predictive model. We then applied a multimodal DL approach with several large language models to both tabular and non-tabular  features for both classification tasks. Just by encoding  textual, categorical and numerical features, we are concatenating a single feature vector that can be fed into a ML model. However, such an approach assumes that combining categorical, text, and numerical features provides complementary information, and that encoding them separately and concatenating them is sufficient to capture this information. There are other approaches for modeling both tabular (i.e., categorical, and numerical) and non tabular data. However, such traditional approaches generally lack the end-to-end pipeline that aggregates different representation schemes of features \cite{chatzis1999multimodal}. 

\begin{figure}[ht]
    \centering
    \includegraphics[width=0.7\textwidth]{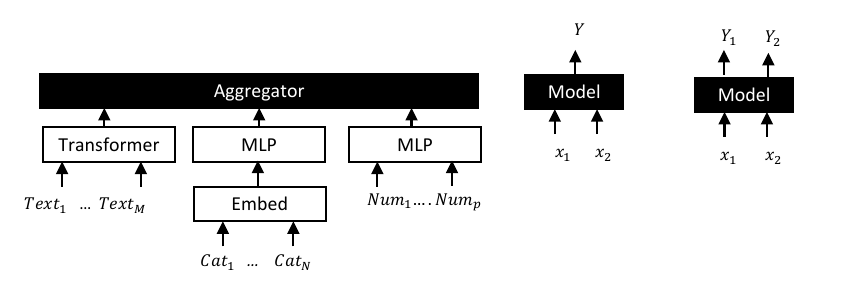}
    \caption{Fuse Late Multimodal Machine Learning}
    \label{fig:fuselate}
\end{figure}

Multimodal ML approaches use neural networks specifically, multi-layer perception (MLP) to process tabular and non-tabular features to produce a set of learned features \cite{morency2022tutorial}. These features, when aggregated, create fully connected layers that are used for both classification tasks ISSQ and ISRs with the right weight given to each modality. The objective of employing a multimodal approach is to take the advantage of the recent trend of using large language models for prediction tasks, encompassing not just textual features but various other feature types as well. We considered an automated supervised learning with multimodals to jointly process our data set \cite{agmultimodaltext}. We used a tabular model together with transformer text architectures. The motivation behind using multimodal neural approach with transformers is that text representations may improve when self-attention is informed by context of numeric/categorical features. 
The multimodal approach is demonstrated in Figure \ref{fig:fuselate}. Individual MLPs are used to process categorical and numerical features, after which they are concatenated with the transformed textual features into a unified input representation \( \boldsymbol{m} = \boldsymbol{x} \| \operatorname{MLP}(\boldsymbol{c}) \| \operatorname{MLP}(\boldsymbol{n}) \). This combination takes as input $\boldsymbol{x}$, representing   textual features generated by a transformer model and  the preprocessed tabular (categorical $(\boldsymbol{c})$ and numerical $(\boldsymbol{n})$) features, and outputs a combined multimodal representation $m$. The parameters generated during this combination are aggregated and fine-tuned using fully connected layers. 
\vspace{-3mm}

 We applied backpropagation to compute token representations for both numerical and categorical features. Furthermore, we incorporated an additional embedding layer to map categorical features into the same $\mathbb{R}^d$ vector space, employing different embedding layers for distinct categorical features. For numerical features, a single hidden-layer MLP was employed to obtain a unified $\mathbb{R}^d$ vector representation. These vectors were then fed into an aggregator encoder to capture the interactions between the embeddings of text tokens, categorical and numerical values. Rather than combining information from different data types early in the multimodel architecture, we processed each data type separately and merged them near the output layer. The approach extracts higher-level features specific to each modality before combining them. Finally, the top vector representations from all three modalities are combined into a single vector using mean/max pooling and concatenation and fed into two dense layers for making predictions.

\subsection{Language Models}

For both classification tasks, we used various large language models to generate text embeddings. One of the most popular models is the Bidirectional Encoder Representations from Transformers (BERT). BERT was designed to pre-train deep bidirectional representations from unlabeled text by jointly conditioning on both left and right context. To make BERT handle a variety of tasks, the input representation should be able to represent both a single sentence and a pair of sentences in token sequences. A ``sequence” refers to the input token sequence to BERT, which may be a set of words, or sentences packed together. The pre-trained BERT model can be fine-tuned with a single additional output layer without substantial task specific architecture modifications. We used BERT to produce fixed sized sentence embeddings for non-tabular features. BERT adds a pooling operation to the output to derive a fixed sized output vectors that can be used for tasks such as classification \cite{devlin2018bert}. 

Compared to BERT, RoBERTa is pretrained model that utilizes a masked language modeling objective with dynamic masking. It was trained using a training technique called byte-level byte pair encoding (BPE), which replaces words with sub-word units. RoBERTa was trained using larger batches and for a longer period compared to BERT \cite{liu2019roberta}. 
We utilized other language models such as GOOGLE ELECTRA \cite{clark2020electra}. In the medical field, G-ELECTRA has demonstrated a notable advantage over BERT in the context of multi-hop medical question answering tasks \cite{li2022easy}.

We used medical language models such as Microsoft BiomedNLP PubMedBERT model (M-BIOMEDNLP) \cite{gu2021domain}, which is based on BERT and was trained on biomedical literature, including PubMed abstracts and full-text articles. The model is fine tuned on tasks such as question answering and named entity recognition. We also tested our approach using the GPT-based transformer Microsoft’s BioGPT \cite{luo2022biogpt}, which is a domain-specific chatbot based on a transformer language model that is used for answering biomedical questions. This model was trained on a massive data set of biomedical literature, including PubMed abstracts, clinical trials, and scientific papers. The model has 1.5 billion parameters, making it one of the largest biomedical language models.
\vspace{-2mm}

\section{Results}
\vspace{-2mm}
This section discusses the results of our experiments. All tests were conducted using an NVIDIA A100 GPU with 80 GBs of VRAM. For each experiment, an 80/20 split was used for training and testing sets.

\begin{table}[ht]
\footnotesize
\centering
\caption{Hyperparameters of Language Models}
\label{tab:HP}
\begin{tabular}{|l|l|}
\hline

\hline
\textbf{Model Hyperparameters} & The size of the embedding layer (768). \\
                                       & Activation function for hidden layers (GELU). \\
                                       & Dropout probability for hidden layers (0.2). \\
                                       & Size of the hidden layers (768). \\
                                       & Range for initializing weights (0.02). \\
                                       & Size of the intermediate layer (3072). \\
                                       & Epsilon value for layer normalization (1e-12). \\
                                       & Maximum position embeddings allowed (512). \\
                                       & Number of attention heads (12). \\
                                       & Number of hidden layers (12). \\
                                       & Type of position embeddings (absolute). \\
                                       & Size of the token type vocabulary (2). \\
                                       & Size of the vocabulary (30522). \\
\hline
\textbf{Other Model Configurations} & Type of the model (ELECTRA, BERT, etc.). \\
                                          & Token ID for padding (0). \\
                                          & Activation function for summary layers (GELU). \\
                                          & Type of summary layer (first). \\
                                          & Whether to use projection for summary (yes). \\
                                          & transformers\_version: 4.26.1 \\
                                          & Whether to use cache during processing (yes). \\
\hline
\end{tabular}
\end{table}

\subsection{Neural Network Optimization and Architectures}

We employed a single hidden layer Multi-Layer Perceptron (MLP) for feature encoding. This MLP consists of a bottleneck layer and implements layer normalization techniques. The activation function used within the MLP layers is a leaky ReLU with a slope parameter of 0.1. Additionally, we utilized the GELU activation function, as outlined in the methodology presented in \cite{devlin2018bert}. The Feedforward Network (FFN) of the Transformer model is configured with an embedding layer size of 768, 12 attention heads, and a hidden layer dimension of 768.

After encoding, all categorical features are combined and subsequently passed through a dedicated basic MLP layer. Similarly, numeric features go through a similar procedure where they are combined and then fed into their own MLP layer. These MLP layers  consist of 128 bottleneck units, and their output dimensions are customized to match the dimensions of the token embeddings in the pretrained Transformer model.
Additionally, the architectural design includes an embedding layer with 32 units, which is subsequently processed by another basic MLP layer with 64 bottleneck units. To ensure consistency, the output dimensions of this MLP are configured to align with the token embedding size of pretrained NLP models, such as ELECTRA. The hyperparameters of the language models used in our experiments are detailed in Table \ref{tab:HP}. The initial learning rate is set at 0.02, with a maximum learning rate of $5 \times 10^{-5}$, and a warmup factor of 0.1. We employ a batch size of 128, incorporating a weight decay of $10^{-4}$, and utilizing the AdamW optimizer. The training process lasts for 40 epochs, with early stopping based on validation performance.

\begin{table}[t!]
\centering
\caption{Distribution of questions and responses in the labeled data}
\label{tab:label_dist}
\footnotesize
\centering
\begin{tabular}{lcc}
\hline
\textbf{Medical Condition} & \textbf{ISSQ} & \textbf{ISR} \\ \hline
Cancer         & 526 out of 600 (88\%) & 385 out of 554 (69\%) \\
Diabetes       & 176 out of 196 (90\%)   & 144 out of 178 (81\%)   \\
Cardiovascular & 270 out of 291 (93\%) & 225 out of 277 (81\%) \\
Neurological   & 649 out of 860 (75\%) & 477 out of 664 (72\%) \\ \hline
\textbf{Sum}   & \textbf{1621 out of 1947 (83\%)} & \textbf{1231 out of 1673 (74\%)} \\ \hline
\end{tabular}
\end{table}

Table \ref{tab:label_dist}, shows the distribution of questions and responses in our labeled data set. In total 18 features were used in our experiments: 2 non-tabular (textual) features, 7 emotional features, 3 text metadata features, 3 post features, and 3 user features. It is worth mentioning that text features were transformed based on the model used (i.e., TF-IDF, or language model embeddings). As such, the final dimensionality of the data depends on the dimensions of the embedding vectors. 

We conducted two classification task: first, to classify questions into ISSQs or non-ISSQs and second, to classify responses to ISRs and non-ISRs. Table \ref{tab:label_dist} shows the distribution of these classes in our labeled data. 



Results of predicting ISSQ is shown in table \ref{tab:QC}. The highest performance values are attained using the multimodal approach with ROBERTA and B-ROBERA language models. Other models such as the ensemble multimodal approach with G-Electra language model yields satisfactory results. This approach utilized a straightforward aggregation strategy that takes a weighted average of the predictions from the language model and various tabular models. Here, the language model and other tabular models are independently trained using a common training/validation split. Subsequently, ensemble selection is used as a greedy forward-selection strategy to fit aggregation weights over all models’ predictions on the held-out validation data. It is also noticed that ROBERTA yields significant advantage in terms of both Accuracy and F1 score. In general, our results indicate that multimodal approaches outperform conventional classifiers, yielding superior results.

\begin{table}[b!]
\caption{Model comparison for informational support question prediction task}
\label{tab:QC}
\centering
\footnotesize
\begin{tabular}{lllllll}
\hline
\textbf{ML Approach} & \textbf{Model} & \textbf{ACC} & \textbf{AUC} & \textbf{F1} & \textbf{Precision} & \textbf{Recall} \\ \hline
Multimodal Fuse Late& ROBERTA& \textbf{0.953}& \textbf{0.965}& \textbf{0.972}& 0.964& 0.979
\\
Multimodal Fuse Late & B-ROBERTA& 0.950& 0.960& 0.970& 0.956& 0.985
\\
Weighted Ensemble-L2& G-ELECTRA& 0.948& 0.949& 0.969& 0.953& 0.985
\\
Multimodal Fuse Late & BERT& 0.945& 0.952& 0.967& 0.950& 0.985
\\
Multimodal Fuse Late & M-BIOMEDNLP& 0.943& 0.962& 0.965& 0.961& 0.970
\\
Multimodal Fuse Late & G-ELECTRA& 0.923& 0.955& 0.955& 0.921& \textbf{0.991}
\\
Multimodal Fuse Late &  DMIS-BIOBERT& 0.923& 0.955& 0.955& 0.921& \textbf{0.991}
\\
Multimodal Fuse Late & B-ELECTRA& 0.920& 0.935& 0.951& 0.960& 0.943\\
Gradient Boosting     & TFIDF        & 0.904          & 0.804          & 0.899          & 0.987          & 0.941          \\
Random Forest         & NA           & 0.894          & 0.803          & 0.901          & 0.969          & 0.934          \\
Random Forest         & TFIDF        & 0.860          & 0.695          & 0.849          & \textbf{0.996} & 0.916          \\
Gradient Boosting     & NA           & 0.856          & 0.741          & 0.874          & 0.951          & 0.911          \\
Logistic             & NA           & 0.849          & 0.683          & 0.845          & 0.987          & 0.910          \\
KNN                  & NA           & 0.829          & 0.702          & 0.858          & 0.934          & 0.894          \\
KNN                  & TFIDF        & 0.829          & 0.702          & 0.858          & 0.934          & 0.894          \\
SVM                  & NA           & 0.825          & 0.646          & 0.830          & 0.973          & 0.896          \\
SVM                  & TFIDF        & 0.825          & 0.646          & 0.830          & 0.973          & 0.896          \\
Logistic             & TFIDF        & 0.822          & 0.617          & 0.818          & 0.991          & 0.941   \\
 \hline
\end{tabular}
\end{table}

\begin{table}[h]
\caption{Model comparison for informational support response prediction task}
\label{tab:AC}
\centering
\footnotesize
\begin{tabular}{lllllll}
\hline
\textbf{ML Approach} & \textbf{Model} & \textbf{ACC} & \textbf{AUC} & \textbf{F1} & \textbf{Precision} & \textbf{Recall} \\ \hline
Multimodal Fuse Late & G-ELECTRA    & \textbf{0.907} & 0.929          & \textbf{0.940} & 0.907          & \textbf{0.976} \\
Multimodal Fuse Late & B-ROBERTA    & 0.901          & 0.926          & 0.935          & 0.926          & 0.944          \\
Multimodal Fuse Late & ROBERTA      & 0.895          & 0.925          & 0.932          & 0.912          & 0.952          \\
Multimodal Fuse Late & M-BIOMEDNLP  & 0.886          & 0.927          & 0.926          & 0.902          & 0.952          \\
Multimodal Fuse Late & BERT         & 0.880          & 0.935          & 0.919          & 0.931          & 0.908          \\
Multimodal Fuse Late & DMIS-BIOBERT & 0.873          & \textbf{0.944} & 0.901          & 0.876          & 0.928          \\
Multimodal Fuse Late & B-ELECTRA    & 0.858          & 0.913          & 0.908          & 0.895          & 0.920          \\
Weighted Ensemble-L2  & G-ELECTRA    & 0.849          & 0.895          & 0.905          & 0.868          & 0.944          \\
Gradient Boosting     & TFIDF        & 0.822          & 0.678          & 0.831          & 0.960          & 0.891          \\
Random Forest         & NA           & 0.820          & 0.701          & 0.844          & 0.934          & 0.887          \\
Random Forest         & TFIDF        & 0.810          & 0.636          & 0.811          & 0.976          & 0.886          \\
Gradient Boosting     & NA           & 0.801          & 0.740          & 0.874          & 0.860          & 0.867          \\
Logistic             & TFIDF        & 0.789          & 0.576          & 0.785          & 0.992          & 0.877          \\
Logistic             & NA           & 0.789          & 0.600          & 0.795          & 0.971          & 0.874          \\
SVM                  & TFIDF        & 0.756          & 0.500          & 0.756          & \textbf{1.000} & 0.861          \\
SVM                  & NA           & 0.755          & 0.500          & 0.755          & \textbf{1.000}          & 0.860          \\
KNN                  & TFIDF        & 0.744          & 0.559          & 0.780          & 0.920          & 0.845          \\
KNN                  & NA           & 0.739          & 0.554          & 0.777          & 0.918          & 0.842         \\ 
 
 \hline
\end{tabular}
\end{table}

Using TF-IDF features with tabular features when applying baseline classifiers didn't significantly enhance classification metrics. For lazy learners SVM and KNN, we did not observe significant differences when applying both types of features. Since both SVM and KNN can handle non-linear decision boundaries, this makes them suitable for classification problems containing numerical features. In other words, using textual features does not add much to the effectiveness of these classifiers. Finally, it is noticed that using both types of features in ensemble models such as gradient boosting leads to better classification results compared to using tabular features only.

Results of identifying informational support responses (ISRs) to medical questions are shown in Table \ref{tab:AC}. It is worth mentioning that for the classification task of ISRs, both language models and multimodal approaches consistently generate the most precise predictions. On average, different classification metrics tend to produce lower scores for predicting ISRs compared to ISSQs suggesting that identifying ISRs is a more challenging task. Interestingly, the addition of textual features in our study did not improve the performance of baseline ML classifiers such as KNN, SVM, and Random Forest, as has been found in other contexts \cite{Marshan2023Comparing}.

    \begin{table}[h]
    \centering
    \caption{Ablation Analysis}
    \label{tab:ablation}
    \resizebox{\textwidth}{!}{ %
    \begin{tabular}{lccccc}
    \hline
    \textbf{FEATURES}                   & \textbf{Accuracy} & \textbf{AUC}   & F1    & Precision      & Recall         \\ \hline
    {[}Full Model{]} = ELECTRA + Emotions + Numeric Text + Post + User & \textbf{0.907} & 0.929          & \textbf{0.940} & 0.907          & 0.976          \\
    {[}Text Only{]} = ELECTRA                                          & 0.892          & 0.918          & 0.931          & 0.897          & 0.968          \\
    {[}No Emotion{]} = ELECTRA + Numeric Text + Post + User            & 0.898          & 0.923          & 0.933          & \textbf{0.925} & 0.940          \\
    {[}No User{]} = ELECTRA + Emotions + Numeric Text + Post           & 0.898          & 0.932          & 0.936          & 0.892          & \textbf{0.984} \\
    {[}No Numeric{]} = ELECTRA + Emotions + Post + User                & 0.901          & 0.934          & 0.937          & 0.898          & 0.980          \\
    {[}No Post{]} = ELECTRA + Emotions + Numeric Text + User           & 0.895          & \textbf{0.940} & 0.931          & 0.922          & 0.940            \\ \hline
    \end{tabular}
    }
    \end{table}

\begin{table}[b!]
\caption{ Model transferability results for informational support response prediction task}
\label{tab:transfer}
\centering
\footnotesize
\begin{tabular}{lllllll}
\hline
\textbf{ML Approach} & \textbf{Model} & \textbf{ACC} & \textbf{AUC} & \textbf{F1} & \textbf{Precision} & \textbf{Recall} \\ \hline
Multimodal Fuse Late & ROBERTA      & \textbf{0.876} & 0.929          & \textbf{0.896} & 0.923      & \textbf{0.870}          \\
Multimodal Fuse Late & G-ELECTRA    & 0.858          & 0.915          & 0.882          & 0.896              & \textbf{0.870}          \\
Multimodal Fuse Late & BERT         & 0.823 & \textbf{0.935} & 0.846 & 0.902              & 0.797 \\
Multimodal Fuse Late & M-BIOMEDNLP  & 0.814          & 0.923          & 0.835          & 0.914              & 0.768          \\
Multimodal Fuse Late & B-ROBERTA    & 0.805          & \ 0.922 & 0.817          & 0.961              & 0.710          \\
Multimodal Fuse Late & DMIS-BIOBERT & 0.77           & 0.897          & 0.780          & 0.939              & 0.667          \\
Random Forest         & TFIDF        & 0.761          & 0.763          & 0.839          & 0.754              & 0.794          \\
Weighted Ensemble-L2  & G-ELECTRA    & 0.752          & 0.847          & 0.770          & 0.887              & 0.681          \\
Logistic             & TFIDF        & 0.735          & 0.700          & 0.747          & 0.855              & 0.797          \\
Multimodal Fuse Late & B-ELECTRA    & 0.726          & 0.842          & 0.735          & 0.896              & 0.623          \\
Logistic             & NA           & 0.664          & 0.585          & 0.657          & 0.942              & 0.774          \\
Gradient Boosting     & TFIDF        & 0.655          & 0.664          & 0.768          & 0.623              & 0.688          \\
Random Forest         & NA           & 0.637          & 0.682          & 0.868          & 0.478              & 0.617          \\
SVM                  & TFIDF        & 0.611          & 0.500          & 0.611          & \textbf{1.000}     & 0.758          \\
SVM                  & NA           & 0.611          & 0.500          & 0.611          & \textbf{1.000}     & 0.758          \\
Gradient Boosting     & NA           & 0.593          & 0.601          & 0.709          & 0.565              & 0.629          \\
KNN                  & TFIDF        & 0.566          & 0.476          & 0.598          & 0.884              & 0.713          \\
KNN                  & NA           & 0.566          & 0.476          & 0.598          & 0.884              & 0.713         \\
 \hline
\end{tabular}
\end{table}
 \vspace{-4mm}
 
\subsubsection{Ablation Study}

Ablation study describes the process of removing parts of a neural network to gain a better understanding of the network’s behavior \cite{girshick2014rich}. We conduct ablation analysis to understand how different groups of features contribute to the overall performance of our best performing DL classifier. As shown in Table \ref{tab:ablation}. The results suggest that our full model with all the 5 sets of features outperforms the reduced model variations in terms of accuracy and F\-1 score. This highlights the suitability of our feature selection approach.

\subsubsection{Model Transferability}

The summary shown in table \ref{tab:transfer} suggests that our multimodal DL approach can be applied to predict ISRs in data sets that were not included in the training of the model. In other words, the model can be transferred to new sub-communities (medical conditions) to make predictions on ISRs with good accuracy. Using the same labeling approach discussed before, we labeled 113 responses from the pregnancy sub-community. We then used the labeled data to test the transferability of our model to this data and its applicability to a different medical condition. Testing environment, tuning parameters, and experiment settings were similar to our previous experiment. The results shown in Table \ref{tab:transfer} validate that our model is transferable to other domains, yielding a satisfactory effectiveness in terms of Accuracy, F1, and Precision. The highest accuracy achieved on the domain shift data is 87.6\%, which is quite close to the accuracy of the original model. Model transferability results also validate that the proposed multimodal approaches and the applied language models are effective in identifying ISRs. 

\subsection{Explanation of Model Features}
We use ``SHapley Additive exPlanations” (SHAP) to interpret the results of our best performing models. SHAP is an explanation method that builds on the concept of Shapely values \cite{lundberg2017unified}. As a coalitional game theory concept, Shapely values provide an approach to fairly distribute gains among players who have made unequal contributions to the outcome of a game \cite{shapley1953value}. Similarly, SHAP assumes the input features of a model as a team of players working together to produce an outcome (i.e., the prediction) and computes the average marginal contribution of each feature accordingly. These computations can be made both for the entire sample (global interpretability) and for individual cases (local interpretability) \cite{zhou2021inferring}\cite{Delen2024Explainable}.

\subsubsection{Global Interpretation}
\begin{figure}[h]
    \begin{subfigure}{0.49\textwidth}
    \centering
    \includegraphics[width=\textwidth]{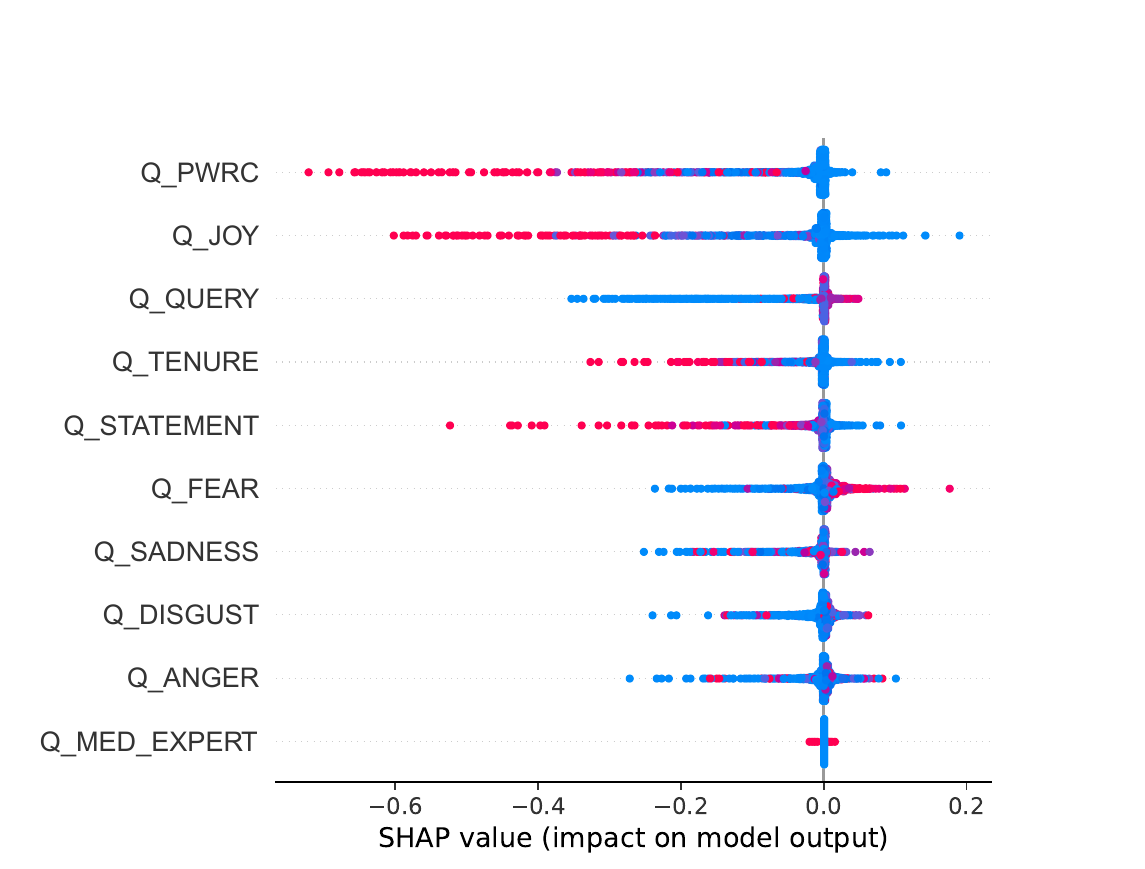}
    \caption{ISSQ Features}
    \label{issqfeatures}
    \end{subfigure}
    \hfill
    \begin{subfigure}{0.49\textwidth}
    \centering
    \includegraphics[width=\textwidth]{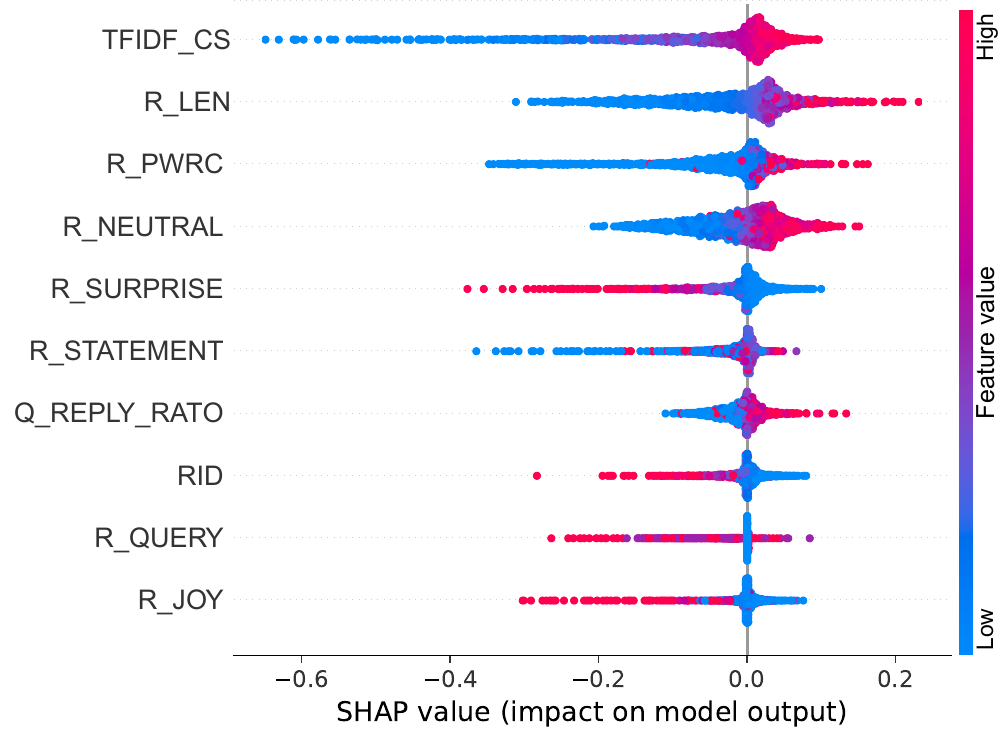}
    \caption{ISR Features}
    \label{isrfeatures}
    \end{subfigure}
    \caption{Feature Importance and Effect Plots}
    \label{fig:feature_importance}
\end{figure}

The summary plots shown in Figure \ref{fig:feature_importance} describe both the importance and the effect of each feature on the two outcome variables ISSQ (Figure \ref{issqfeatures}) and ISR (Figure \ref{isrfeatures}) for the entire sample.

\begin{figure}[b!]
    \centering
    \begin{subfigure}{\textwidth}
    \includegraphics[width=\linewidth]{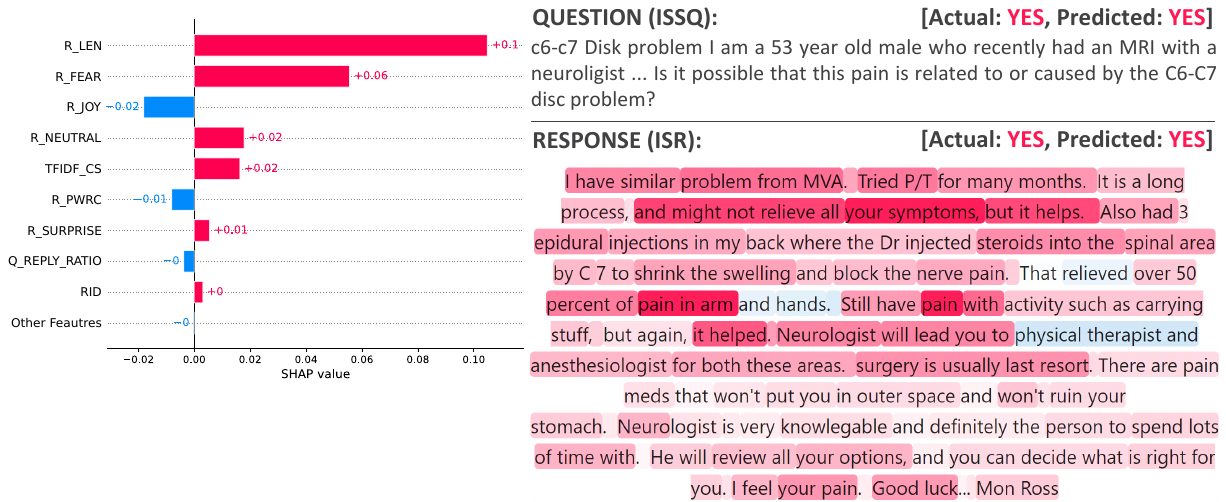}
    \subcaption{Correct Informational Support Response}
    \label{fig:InformationalSupportResponse}
    \end{subfigure}
    \caption{Example Texts}
\end{figure}

\begin{figure}[t!]\ContinuedFloat
    \centering
        \begin{subfigure}{\textwidth}
    \includegraphics[width=\linewidth]{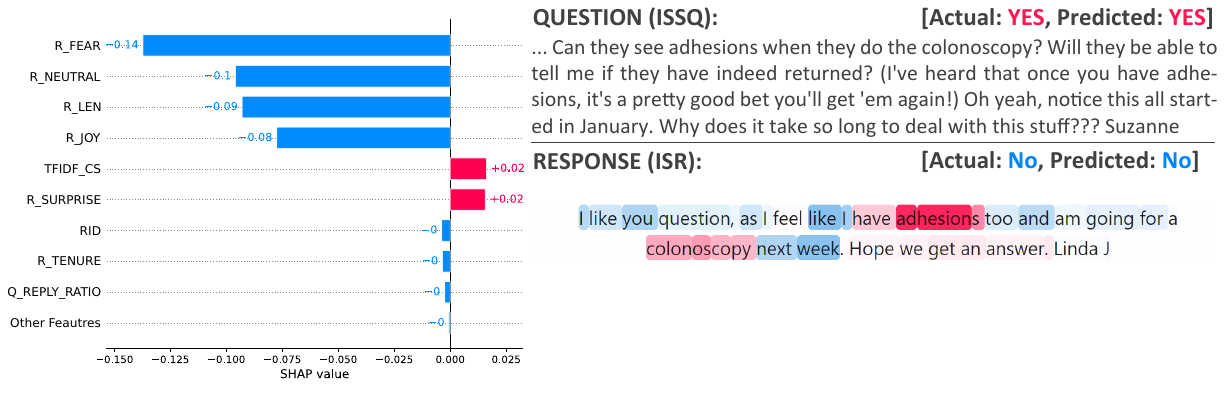}
    \subcaption{Correct Non-Informative Support Response}
    \label{fig:NonInformationalSupportResponse}
    \end{subfigure}
    \begin{subfigure}{\textwidth}
    \includegraphics[width=\linewidth]{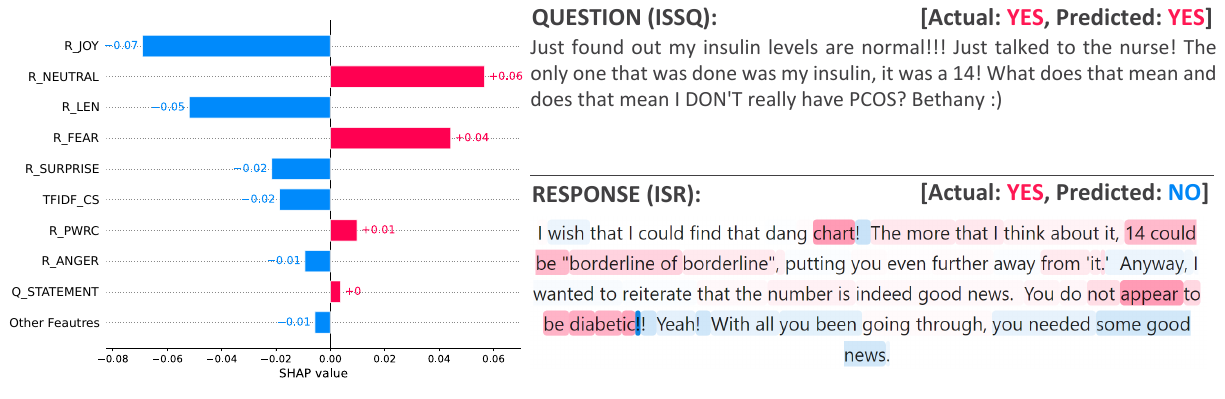}
    \subcaption{False Negative}
    \label{fig:FalseNegative}
    \end{subfigure}
    \begin{subfigure}{\textwidth}
    \includegraphics[width=\linewidth]{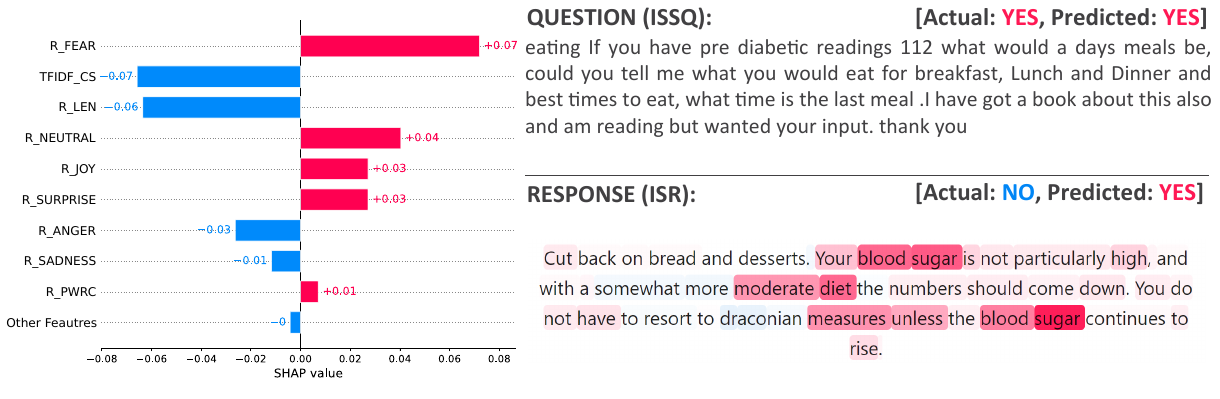}
    \subcaption{False Positive}
    \label{fig:FalsePositive}
    \vspace{-2mm}
    \end{subfigure}
    \caption{Example Texts (cont.)}
    \vspace{-2mm}
    \label{fig:ExampleTexts}
\end{figure}

Features are sorted in Figure \ref{fig:feature_importance} by descending order of importance to the model. The value of 0.0 on the X-axis indicates where the feature is neither contributing to informational support nor non-informational support. As the values of the X-axis move positive from zero, the features contribute more toward informational support, and as the move negative from zero, the features contribute more toward non-informational support. Colors indicate the feature value from low (blue) to high (red). Also, the red-purple-blue gradient provides a sense of the directionality of the impact of each feature. The thickness of lines show the density of observations at each value.

User's platform-wide response count (Q\_PWRC) is identified as the most important predictor in Figure \ref{issqfeatures}, with the widest range of values. Higher values of Q\_PWRC (shown in red) are linked to the lower (non-ISSQ) class. A similar pattern is observed with the user's platform tenure (Q\_TENURE). The second most important predictor is Q\_JOY, where higher values are also associated with lower probabilities of ISSQ. Except for fear (Q\_FEAR), the increase in other negative emotions such as disgust, fear, and sadness are linked with the ISSQ class. The third feature in the list is the number of running queries in the question post (Q\_QUERY), and an increase in the number of query sentences is linked with an increased likelihood of the ISSQ class.

Figure \ref{isrfeatures} on the right highlights the key factors in predicting ISR. It is evident that question response similarity (TFIDF\_CS) exhibits the highest positive relationship with the ISR class followed by text length (R\_LEN) and the user’s platform-wide response count (R\_PWRC). Additionally, the absence of emotional valence (R\_NEUTRAL) is positively linked to the ISR class, while an increase in surprise emotion (R\_SURPRISE) is associated with the non-ISR class.

\subsubsection{Local Interpretation}
Figure \ref{fig:ExampleTexts} shows several example question-response pairs from the data. The words in the responses texts are colored blue for non-informational support words and red for informational support words, as identified by SHAP. The tabular features on the left are sorted by descending level of SHAP values, as detected in the response text on the right. The length of the bars represents the relative strength of the feature in the text. The order of the features changes for each example, depending on the particular text. Figure \ref{fig:InformationalSupportResponse} shows a medical question being answered with a high level of informational support. The model correctly determined it was informational support. Note that even though the response was strongly classified as informational support, the SHAP plot shows that emotions were detected in the response. Some emotions contributed to informativeness and R\_JOY contributed against it. Figure \ref{fig:NonInformationalSupportResponse} shows a information seeking question being answered with an emotional support response. This response contained no informational support and was correctly predicted as a non-informational response by the model.

Figure \ref{fig:FalseNegative} shows where the model was incorrect in its assessment. In the response, the user is not only sharing informational support in interpreting test results, but is also sharing much emotional support in the last sentence. The model picked up on both of these, but the emotional support at the end was stronger and the response was incorrectly predicted. Figure \ref{fig:FalsePositive} was predicted to be informational support by the model, but this response did not answer the question the user had about diet. The response has several factual elements to it, but they do not combine to answer the question. 

\subsubsection{The Effect of ISR on Perceived Helpfulness}

\begin{figure}[tp!]
    \centering    
    \includegraphics[width=0.7\textwidth]{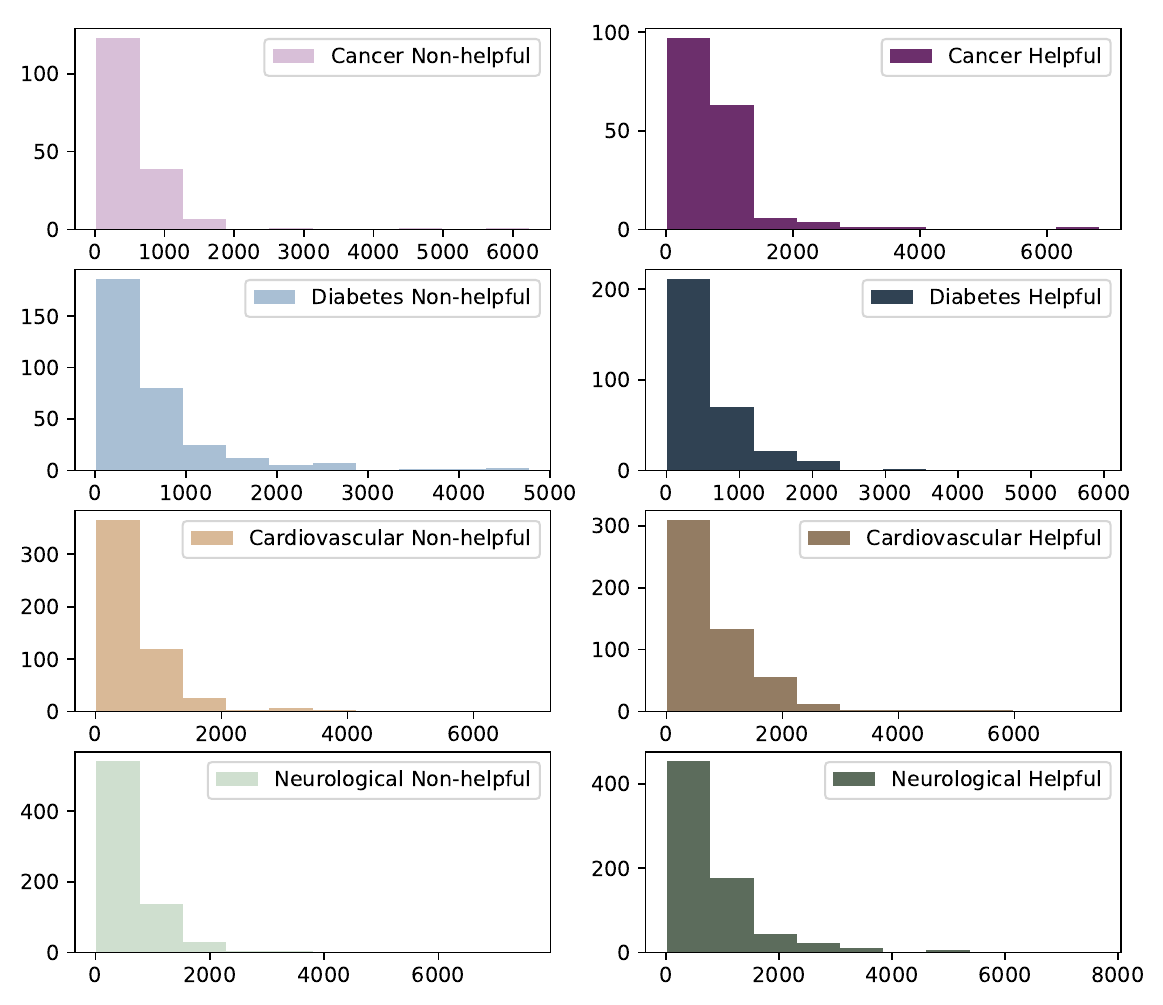}
    \caption{The Distribution of the Range of Characters}
    \vspace{-3mm}
    \label{fig:paired_samples}
\end{figure}

To answer research question 3, ``Does informational support responses predict the helpfulness of a response?” we examined 631,522 responses related to the four different medical conditions in this study (cancer, diabetes, cardiovascular, and neurological conditions). Of these, only 1,733 responses (0.3\%) were identified as helpful by users of the OHC. Given the highly imbalanced nature of the data set, we undersampled responses that lacked the ``helpful” designation. We used medical condition type, and character length distribution as our matching criteria. These distributions are visualized in Figure \ref{fig:paired_samples}. Using our best performing models, we first predicted ISSQ and subsequently determined the ISR values for each question-response pair. Table \ref{tab:isr_hlp_freq} presents the frequency of observations in each class. In the sample 1032 (34\%) ISRs were not marked as helpful by OHC users. 

\begin{table}[h]
\small
\centering
\caption{The Frequency of Non-ISR, and ISR per Non-helpful and Helpful Responses}
\label{tab:isr_hlp_freq}
\begin{tabular}{cccc}
\multicolumn{1}{l}{} & \textbf{} & \multicolumn{2}{c}{\textbf{HELPFUL}} \\ \cline{3-4} 
\textbf{} & \multicolumn{1}{c|}{} & \multicolumn{1}{c|}{0} & \multicolumn{1}{c|}{1} \\ \cline{2-4} 
\multicolumn{1}{c|}{{\rotatebox[origin=r]{90}{\textbf{ISR}}}} & \multicolumn{1}{c|}{0} & \multicolumn{1}{c|}{436} & \multicolumn{1}{c|}{353} \\[-10pt] \cline{2-4} 
\multicolumn{1}{c|}{} & \multicolumn{1}{c|}{1} & \multicolumn{1}{c|}{1032} & \multicolumn{1}{c|}{1246} \\[0pt] \cline{2-4} 
\end{tabular}
\end{table}

Out of the 1,735 responses lacking the ``helpful” designation, 1,032 (59\%) were categorized as ISR. However, among the 1,733 helpful responses, 1,246 (72\%) were predicted as ISR. Subsequently, we employed logistic regression to investigate the impact of ISR on response helpfulness while accounting for potential confounding factors, including text length, type of medical condition, responder's platform-wide response count, and their medical expertise. The findings are presented in Table \ref{tab:helpfulness} and the VIF values are presented in Appendix C. As shown in the table, the odds of a response being marked as helpful is 32\% higher when that response offers informational support (p $<$ 0.01). Moreover, the length of the response positively affects the likelihood of perceived helpfulness such that for each additional unit increase in the number of text characters, the log-odds of a response receiving a helpful designation is 0.03\% higher while holding all other variables constant. Other variables including the type of medical condition, platform-wide response count, and medical expertise of the responder were not found to be significant predictors of perceived helpfulness.

\begin{table}[h]
\centering
\caption{The Effect of ISR on Helpfulness}
\label{tab:helpfulness}
 \resizebox{.6\textwidth}{!}{%
\begin{tabular}{lcc}
\hline
\textbf{DV: HEPLFUL\_DUM} & \textbf{Logit} & \textbf{Odds Ratio} \\ \hline
ISR & 0.278** (0.09) & 1.3203 \\
R\_LEN & 0.0003*** (0.00) & 1.0003 \\
M\_CONDITION & 0.055 (0.04) & 1.0563 \\
R\_PWRC & 0.000 (0.00) & 1.0000 \\
Q\_MED\_EXPERT & 0.013 (0.32) & 1.0133 \\
Intercept & -0.802* (0.35) & 0.4484 \\ \hline
Number of Obs. & \multicolumn{2}{c}{3067} \\
Log Likelihood & \multicolumn{2}{c}{-2091.93} \\
Chi2(5) & \multicolumn{2}{c}{51.14} \\ \hline
\multicolumn{3}{c}{{\begin{tabular}[c]{@{}c@{}}Notes: Standard errors in parentheses. Results are reported with robust errors. \\ ***p \textless 0.001; **p \textless 0.01; *p \textless 0.05.\end{tabular}}} \\
\multicolumn{3}{c}{} \\
\multicolumn{3}{c}{}
\end{tabular}}
\vspace{-15mm}
\end{table}

\section{Discussion and Contribution}
 \vspace{-2mm}

The goals of this study were to develop advanced techniques to classify informational support questions and responses accurately and reliably by using question-response pairs, to investigate the role of emotions in determining the informational/non-informational support types in OHCs, and to explore the relationship between the provision of informational support in a response and its helpfulness across a variety of medical conditions.

By considering question-response pairs, we were able to achieve significant results in predicting ISSQ and ISR classes, with an accuracy of 95.3\% and 90.7\% respectively. Additionally, model transferability results were favorable, as we were able to predict ISR in an unrelated context with an accuracy of 86.7\%. This result suggests that the models we developed have the potential to be applied to other medical conditions, thus, providing further insight into the process of solicitation and provision of informational support by OHC users.

Furthermore, our findings indicate that both the request for and provision of informational support is more entangled with emotions than previously acknowledged in the literature. Although the neutral tone is a major predictor of the ISR class, it is important to remember that all emotion scores, including the neutral score, are continuous values ranging from 0 to 1. Thus, the fact that R\_NEUTRAL positively contributes to ISR does not imply that all neutral responses are informational. Analyzing the R\_NEUTRAL distribution in the Appendix Figure \ref{fig:r_emo_dist} reveals that when a text is emotionally charged and has a neutral score close to zero (for example \textit{``This is great news, Chris. I am happy for her. Have a safe trip. Marie”}), the response is unlikely to contain information. However, responses with higher degrees of neutrality (neutral scores greater than 0.5) are more likely to be informational as neutral responses are less susceptible to biases \cite{kahneman2011thinking, johnson2008we}.

These findings challenge the traditional way of viewing emotional and informational support as mutually exclusive constructs. Emotional support often complements informational support, and emotions can sometimes serve as tools for improving the effectiveness of information. For instance, the use of sympathy and fear in a message can enhance the reception and the persuasiveness of advice \cite{Feng2010, das2003fear}. This could mean that a certain level of emotion may be needed to convey important subtext in a message. 
The relationship of fear and joy in ISRs is also worth further examination. As shown in Appendix Figure \ref{fig:fear_joy}, high values of joy are correlated with low values of fear, and conversely, low levels of joy are correlated with high levels of fear. While low fear and high joy responses are mainly non-informational, high fear and low joy responses often predict the ISR class. These findings highlight the importance of considering emotions when studying informational support in OHCs and suggest that emotional and cognitive factors should be integrated in models that explicate informational support. Strong positive emotions in the questions may indicate that the questioner is seeking emotional support. While strong negative emotions both in questions and responses are often indicative of request or provision of informational support. Therefore, OHCs may need to use different strategies to elicit and provide informational support depending on the emotional context of the communication.

The most important tabular predictor for the ISSQ classification task is the platform-wide response count of the questioner (Q\_PWRC). This indicates that members who are actively engaged with the community tend to be more inclined towards seeking emotional/conversational support or they may act as community facilitators who often share news and updates. Similarly, existing members of OHCs are more likely to seek emotional support, whereas newer members tend to seek informational support. This finding indicates that the motivation for seeking support may change over time and that OHCs must tailor their support offerings to meet the changing needs of their members. 

\subsection{Theoretical Contributions}
The present research has both theoretical and practical applications. First, medical conversations in OHCs are under-studied. Past research has investigated emotional support and community aspects of OHCs (e.g., \cite{Zhou2023}), but few have examined informational support in this context. A growing number of patients rely on OHCs to discuss symptoms, diagnosis, treatment options, risks and benefits associated with medications and surgeries, and preventative measures \cite{bhagat2022conceptualizing}. This variety makes informational support the most common type of social support exchanged in these communities \cite{yan2014feeling}. By considering the ISSQs and ISRs in pairs, we were able to create a comprehensive method for studying the dynamics of social support in OHCs. As evidenced by our findings, the use of advanced classification techniques powered by large language models enabled us to develop dependable and transferable models for predicting informational support in OHCs.

Second, and in answer to the second research question, we find that valenced and neutral emotions both play strong roles in determining ISSQs and ISRs. Scholars typically approach informational support from a cognitive perspective. However, certain emotions, like fear, have been linked to ISRs and may enhance information reception, whereas emotions such as joy, surprise, and anger show no clear association with ISRs. More research is needed to fully elucidate the role of emotions in informational support. For example, studies have found that informational support is easier to receive when also paired with sympathetic content \cite{Feng2010}. It may be that a certain level of emotion is needed to convey important subtext in a message. Our approach to understanding informational support could be considered as the first step in understanding information quality \cite{lee2002aimq} in OHCs.

Third, this research also contributes to social support theory by refining what is understood as informational support. We based our definition of what is informational from information theory, which says that information in the OHC context answers the questions posed by the questioner and reduces information asymmetries \cite{Wyer1999, chen2020linguistic}. Traditionally, it has been accepted that referral texts (i.e., informing the questioner they should pose their question to another person, usually a medical doctor in the case of OHCs) constitute a form of informational support. However, we argue that referring questioners to medical professionals does not decrease the information asymmetry between the questioner and responder since these responses do not answer the question(s) posed by the questioner. Instead, such responses are redirects to other sources that, according to the responder, should have the potential to reduce information asymmetries and answer the questioner's questions. A physician would likely be able to answer the question(s) being posed, but the questioner is asking the OHC community members for help, not their physician.  

Our aim in highlighting this aspect is not to discount the value of referrals. If members of an OHC cannot answer the question(s) asked, the questioner should seek other avenues to have their question(s) answered. Physicians are the best source of such information. Instead, we seek to the importance of context and specificity in providing meaningful informational support within OHCs. For example, responders who share detailed accounts of their similar medical experiences may provide questioners with partial answers to the questioner's question(s). Partial answers to questions, while not eliminating information asymmetries, do reduce them, and thus, are considered informational support. Our intention is to encourage a richer and more informative exchange of support within OHCs, recognizing the diverse information needs and expectations of users.

Fourth, we explored the relationship between ISR, helpfulness and helpful votes (measured as whether the ``helpful” thumbs up button was clicked on responses). In the sample, our analysis revealed that the agreement between the helpful votes and our ML method for identifying ISRs was only 55\%. This discrepancy highlights the potential ambiguities inherent in the ``helpful” indicator, which can be influenced by several factors. First, the MedHelp platform does not encourage users to mark responses as helpful when their questions are answered. This results in only 0.8\% of questions with at least three responses having a response marked as helpful. This under utilization of the ``helpful” button significantly reduces its utility as a scientific measure of what responses are helpful. Second, anonymity and lack of any temporal context associated with the ``helpful” votes by simply clicking ``helpful” button allow for arbitrary designations over time, potentially leading to assessments that do not accurately reflect response helpfulness. Third, the helpful button can be appropriated for many things since it is the only reaction option for responses, i.e., MedHelp has no reactions for ``like”, ``funny”, ``sad”, and so on that are common on other social media platforms. Given these limitations of the helpful rating feature on MedHelp and other OHC platforms, future researchers should investigate the alternate ways that helpful is appropriated by users on OHCs and how these appropriations affect the intended functionality of helpful indicators. 

Relying solely on ``helpful" votes as a measure of actual helpfulness can be unreliable, as indicated by past research \cite{mitra2021helpfulness}\cite{sun2019helpfulness}. Studies investigating online reviews have shown that a significant portion of these reviews go unnoticed by users and fail to receive any ``helpful" votes \cite{choi2020empirical}. The present research found a significant relationship between ISR and ``helpful" votes ($\beta$ = 0.278, p $<$ 0.01; see Table \ref{tab:helpfulness}). We believe this relationship would be stronger if the data set had more responses voted ``helpful" by users. Given this relative scarcity of responses voted ``helpful" by users, future researchers could use the method for identifying ISRs in the present research as an alternative to measuring ``helpful" votes.

\subsection{Practical Contribution}
This study offers several practical implications for both OHC platforms and users. First, this study identifies the factors that facilitate or hinder the provision of high-quality informational support in OHCs. This can facilitate the development of effective strategies for promoting healthy behaviors and improving health outcomes among OHC users. OHC platforms can develop blog posts by drawing from several high informational support responses on the same topic. This will allow a single page/resource to contain more provisions of informational support than what would be found in response to individual questions. 

Second, OHC platforms could use the process presented here to develop a recommendation feature. By leveraging the proposed approach, OHCs can design a tool that effectively detects and accentuates the informational support responses. This feature could help filter out non-informational responses, allowing users to quickly locate the information they seek. Such a tool has the potential to streamline users' information retrieval process, reducing the risk of being subjected to information overload or be adversely influenced by emotional support that is not relevant to their needs.

Third, bad advice itself could be somewhat mitigated by OHCs if the OHC platforms were to scan responses for informational support content at the time they are written. The OHC platform could notify a user that their intended response may not be considered informative to the question poster. The OHC platform could then prompt the user with suggestions to improve the quality of the post. For example, if a user were responding to an ISSQ and the platform found the response to be using emotions such as surprise or sadness which are negatively associated with ISRs (see Table \ref{tab:RegressionForISR}), the system could prompt the user to rephrase their post to have less of these emotions and instead adopt a more neutral tone. 

Fourth, through this research's use of question-response pairs, our approach can be used to identify the data necessary to train conversational AI (chatbots) to answer users' medical questions. These chatbots can be trained to identify informational support responses. When a user asks a medical question, the chatbot can determine similar questions that have been posted and search through the responses to those questions and identify the highest scoring informational support response supplied by a user and provide the informational portions of the message, as determined by SHAP, to the chatbot user as an answer. 

Fifth, our study demonstrates the ML model's capacity for transferring knowledge to other medical conditions, enabling precise prediction in new domains with minimal labeled data. To validate this idea, we assess the model's adaptability across various sub-communities, each characterized by a distinct medical condition, and find evidence of its effectiveness in those communities.
\vspace{-2mm}

\section{Limitations and Future Work}
\vspace{-2mm}
As with any research endeavor, this study is not without limitations. First, the use of DL models requires a considerable amount of data for optimal performance. In our study, we annotated a random sample of 2,000 question-response pairs for training and testing and we used a second test set to examine the model’s transferability. While our sample size is substantial compared to prior studies (e.g. \cite{chen2020linguistic, Zhou2023}), developing a large-scale model that can make accurate predictions across various domains may still require larger labeled data sets.

Future research can focus on building extensive labeled data sets to categorize informational and emotional social support, including their subcategories such as advice, teaching, personal experience for informational support and empathy, affirmation, or encouragement for emotional response. By doing so, the request and provision of social support on OHCs can be viewed not as a binary but as a multi-label classification or a regression problem with scores that range from purely informational to entirely emotional.

As discussed in Section 5.1, ``helpful" votes are not a reliable measure as platform users often under-utilize this feature. To investigate the effect of ISR on helpfulness, we needed to work with a balanced data set. Although we took the necessary steps to acquire a randomly selected matched sample, we acknowledge that the number of responses with the helpful vote is substantially lower in real world scenarios. Therefore, the results presented here should be interpreted with this limitation in mind.

Additionally, following our pairwise view of questions and responses, future studies can examine the alignment of emotions in support requests and emotions in support provisions to gain a more profound insight into the interconnected informational and emotional needs of OHC users and develop effective strategies to address these needs.
\vspace{-2mm}

\section{Conclusion}
\vspace{-2mm}
This research sought to help medical information seekers to identify medical responses that provide informational support. Additionally, it aimed to investigate the connection between the provision of information in a response and users' perception of its helpfulness. Driven by existing large language models, including medical language models, we utilized a novel multimodal approach to identify informational requests and responses based on textual, emotive, and other features. We found factors that facilitate or hinder high-quality informational support in OHCs including the number of interactions that a user has with other members on the platform, the questioner's engagement with the post, and the effect of valenced and neutral emotions. We also show that our model can adapt to other domains, and thus it can be used to predict informational responses when it is not trained on those domains. Finally, our research shows that informational responses have a greater probability of being labeled as helpful. More research is still needed on the role of emotions in informational support messages in OHCs and to delve deeper into the connection between information and how it impacts perceived helpfulness.


\section{Declarations}

\subsection{Authors’ Contributions} All authors contributed to the design and implementation of this research, to the data analysis, and to the writing of the manuscript.

\subsection{Ethics Approval and Consent to Participate} Not Applicable

\subsection{Data Availability} The data that support the findings of this study were collected from the public Internet. Datasets are available from the corresponding author upon request.

\subsection{Consent for Publication} The authors give our consent for all relevant information about this research to be published in \textit{Information Systems Frontiers}.

\subsection{Competing Interests} The authors have no financial or proprietary interests in any material discussed in this research.

\label{}




\setlength{\bibsep}{.6pt}
 \bibliographystyle{elsarticle-num-names}
 \bibliography{references.bib}
\clearpage  

\appendix
\section{Statistical Models}
\vspace{-3mm}
\begin{table}[h]
\footnotesize
\centering
\caption{Regression results for predicting ISSQ}
\begin{tabular}{lcc}
\hline
\multicolumn{1}{c}{\textbf{Independent Variables}} & \textbf{(A) Logistic Regression} & \textbf{(B) Probit Regression} \\ \hline
Q\_ANGER & -0.829 (0.58) & -0.432 (0.33) \\
Q\_FEAR & 1.204*** (0.30) & 0.607*** (0.15) \\
Q\_JOY & -7.329*** (1.01) & -4.147*** (0.49) \\
Q\_SADNESS & -0.932** (0.30) & -0.478** (0.16) \\
Q\_QUERY & 0.220*** (0.06) & 0.106*** (0.03) \\
Q\_STATEMENT & -0.040*** (0.01) & -0.023*** (0.00) \\
Q\_PWRC & 0.000*** (0.00) & 0.000*** (0.00) \\
Q\_TENURE & -0.001*** (0.00) & -0.001*** (0.00) \\
Q\_MED\_EXPERT & -0.735 (0.47) & -0.408 (0.27) \\
Intercept & 2.366*** (0.18) & 1.391*** (0.09) \\ \hline
Number of Obs. & 1,996 & 1,996 \\
LR Chi2(9) & 463.15*** & 452.04*** \\
Pseudo R2 & 0.2564 & 0.2503 \\
Log Likelihood & -671.48825 & -677.0454 \\ \hline
\multicolumn{3}{l}{Note: * p $<$ 0.05; ** p $<$ 0.01; *** p $<$ 0.001. Standard errors in parentheses.}
\end{tabular}
\end{table}

\begin{table}[t!]
\footnotesize
\centering
\caption{Regression results for predicting ISR}
\label{tab:RegressionForISR}
\begin{tabular}{lcc}
\hline
\textbf{Independent Variables} & \multicolumn{1}{l}{\textbf{Logistic Regression}} & \multicolumn{1}{l}{\textbf{Probit Regression}} \\ \hline
R\_ANGER & -0.792 (0.70) & -0.401 (0.42) \\
R\_JOY & -1.557** (0.46) & -0.900** (0.27) \\
R\_NEUTRAL & 0.830** (0.27) & 0.485** (0.15) \\
R\_SADNESS & -1.102** (0.31) & -0.636** (0.18) \\
R\_SURPRISE & -2.073*** (0.40) & -1.234*** (0.23) \\
R\_QUERY & -0.302*** (0.06) & -0.171*** (0.03) \\
R\_STATEMENT & 0.095*** (0.02) & 0.051*** (0.01) \\
TFIDF\_CS & 4.555*** (0.42) & 2.666*** (0.24) \\
RID & -0.038* (0.02) & -0.023* (0.01) \\
R\_PWRC & 0.000 (0.00) & 0.000 (0.00) \\
R\_TENURE & 0.000** (0.00) & 0.000** (0.00) \\
R\_MED\_EXPERT & 0.771* (0.34) & 0.401* (0.18) \\
Q\_REPLY\_RATIO & 0.704* (0.36) & 0.401* (0.20) \\
Intercept & -1.632*** (0.31) & -0.921*** (0.18) \\ \hline
Number of Obs. & 1,643 & 1,643 \\
LR Chi2(13) & 422.17*** & 419.72 \\
Pseudo R2 & 0.2216 & 0.2203 \\
Log Likelihood & -741.58092 & -742.80791 \\ \hline
\multicolumn{3}{c}{\begin{tabular}[c]{@{}c@{}}Note: * p \textless 0.05; ** p \textless 0.01; *** p \textless 0.001. Standard errors in parentheses. \\ R\_FEAR was dropped due to high VIF value.\end{tabular}}
\end{tabular}
\end{table}

\begin{table}[!htb]
\footnotesize
    \begin{minipage}{.5\linewidth}
      \caption{VIFs for ISSQ}
      \centering
        \begin{tabular}{lcc}
            \hline
			\textbf{Variable} & \textbf{VIF} & \textbf{1/VIF} \\ \hline
			Q\_FEAR & 1.35 & 0.74 \\
			Q\_SADNESS & 1.24 & 0.81 \\
			Q\_JOY & 1.17 & 0.86 \\
			Q\_PWRC & 1.13 & 0.89 \\
			Q\_TENURE & 1.10 & 0.91 \\
			Q\_STATEMENT & 1.10 & 0.91 \\
			Q\_QUERY & 1.09 & 0.92 \\
			Q\_ANGER & 1.06 & 0.94 \\
			Q\_MED\_EXPERT & 1.00 & 1.00 \\ \hline
			\textbf{Mean VIF} & \multicolumn{2}{c}{\textbf{1.14}} \\ \hline
        \end{tabular}
    \end{minipage}%
    \begin{minipage}{.5\linewidth}
      \centering
        \caption{VIFs for ISR}
        \begin{tabular}{lcc}
		\hline
		\textbf{Variable} & \multicolumn{1}{l}{\textbf{VIF}} & \multicolumn{1}{l}{\textbf{1/VIF}} \\ \hline
		R\_NEUTRAL & 1.47 & 0.68 \\
		R\_SADNESS & 1.32 & 0.76 \\
		R\_SURPRISE & 1.23 & 0.81 \\
		TFIDF\_CS & 1.16 & 0.86 \\
		R\_PWRC & 1.16 & 0.86 \\
		R\_TENURE & 1.15 & 0.87 \\
		R\_JOY & 1.15 & 0.87 \\
		R\_STATEMENT & 1.14 & 0.88 \\
		R\_ANGER & 1.08 & 0.92 \\
		R\_QUERY & 1.08 & 0.93 \\
		Q\_REPLY\_RA$\sim$O & 1.07 & 0.93 \\
		RID & 1.06 & 0.94 \\
		R\_MED\_EXPERT & 1.03 & 0.97 \\ \hline
		\textbf{Mean VIF} & \multicolumn{2}{c}{\textbf{1.16}} \\ \hline
        \end{tabular}
    \end{minipage} 
\end{table}
\vspace{-5mm}

\section{VIF Values for the Helpfulness Analysis}
\vspace{-3mm}
\begin{table}[H]
\footnotesize
\centering
\caption{VIFs for HELPFUL\_DUM}
\begin{tabular}{lll}
\hline
\textbf{Variable} & \textbf{VIF} & \textbf{1/VIF} \\ \hline
ISR & 1.10 & 0.91 \\
RLEN & 1.09 & 0.92 \\
FORUM\_ID & 1.03 & 0.97 \\
R\_PWRC & 1.03 & 0.97 \\
Q\_MED\_EXPERT & 1.00 & 1.00 \\ \hline
\textbf{Mean VIF} & \multicolumn{2}{c}{\textbf{1.05}} \\ \hline
\end{tabular}
\end{table}

\clearpage
\section{Distributions of emotion scores}
\vspace{-3mm}
\begin{figure}[h!]
\centering
\includegraphics[width=0.8\textwidth]{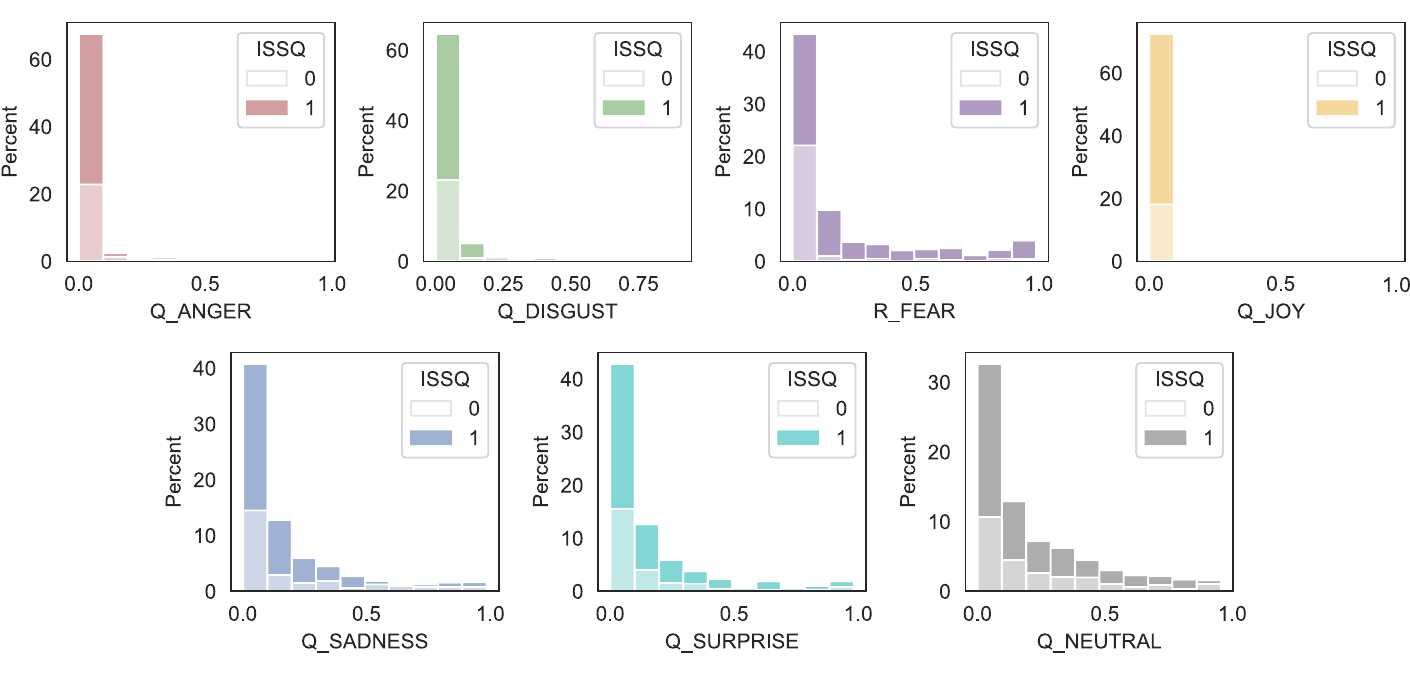}
\caption{Distribution of emotion scores for questions}
\label{fig:q_emo_dist}
\end{figure}
\vspace{-5mm}
\begin{figure}[h!]
\centering
\includegraphics[width=0.8\textwidth]{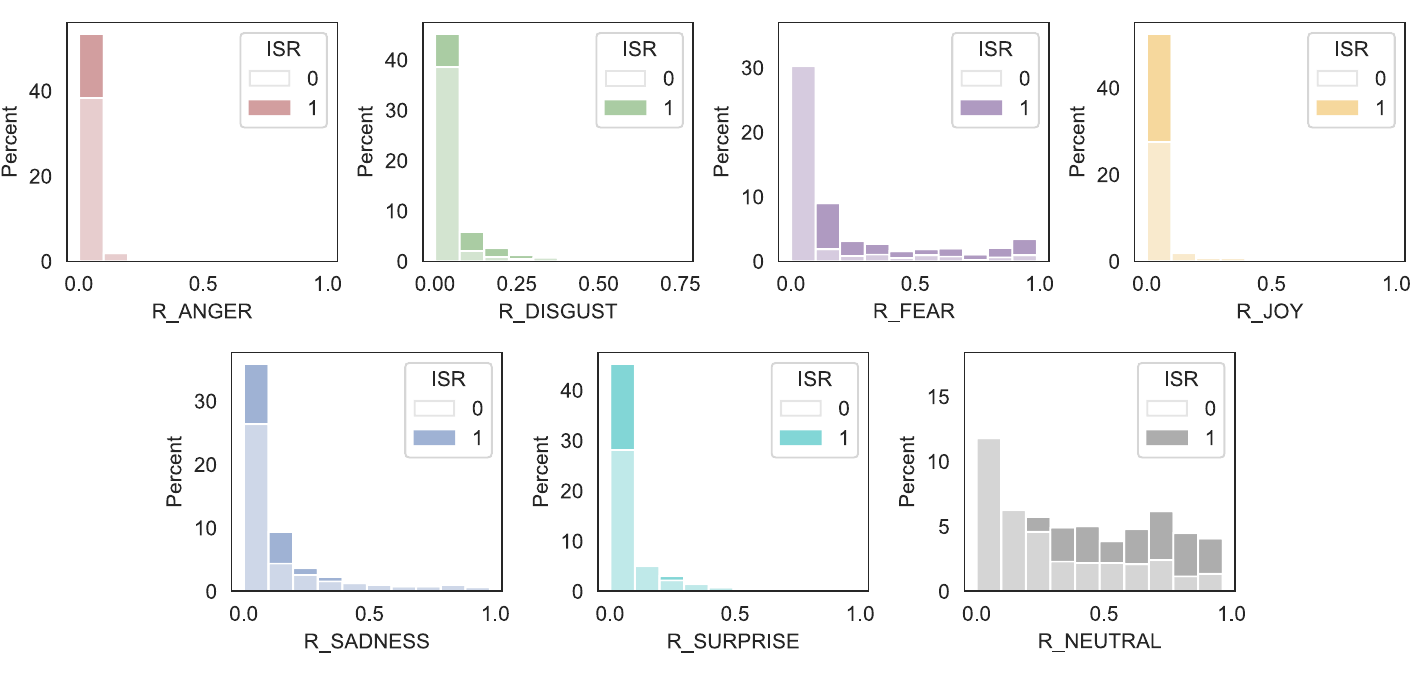}
\caption{Distribution of emotion scores for responses}
\vspace{-3mm}
\label{fig:r_emo_dist}
\end{figure}
\vspace{-5mm}
\clearpage
\begin{figure}[h!]
\centering
\includegraphics[width=.5\columnwidth]
{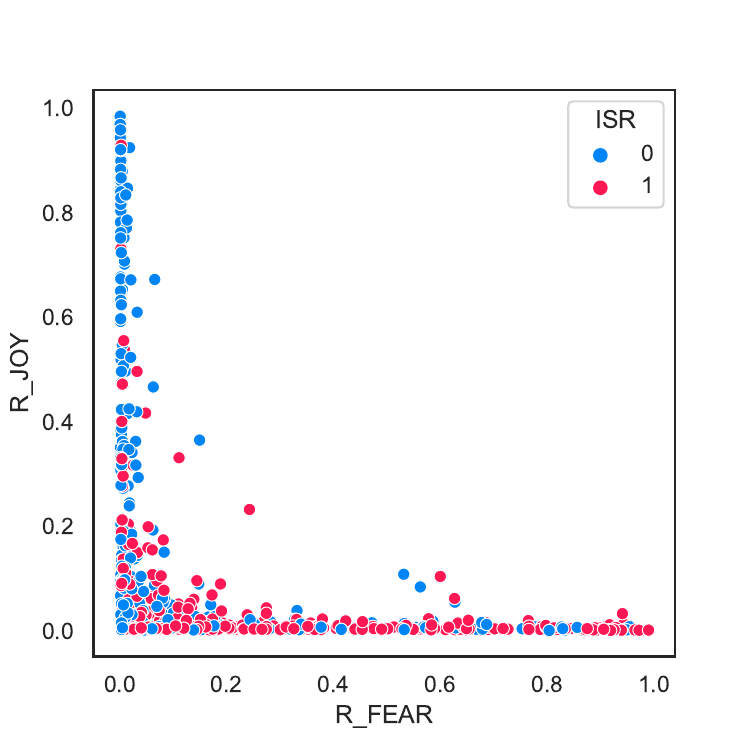}
\caption{R\_FEAR and R\_JOY scatter plot}
\label{fig:fear_joy}
\end{figure}

\end{document}